\documentclass[iicol, table]{sn-jnl}% Default with double column layout
\usepackage{apacite}
\usepackage{soul}
\usepackage{graphicx}%
\usepackage{multirow}%
\usepackage{amsmath,amssymb,amsfonts}%
\usepackage{amsthm}%

\usepackage[title]{appendix}%
\usepackage{xcolor}%
\usepackage{textcomp}%
\usepackage{manyfoot}%
\usepackage{listings}%
\usepackage{tabularx}
\usepackage[table]{xcolor} % For \rowcolors
\usepackage{booktabs}
\usepackage{siunitx}
\theoremstyle{thmstyleone}%
\theoremstyle{thmstyletwo}%

\theoremstyle{thmstylethree}%

\usepackage{mathrsfs}

\begin{document}

\title{RareSpot+: A Benchmark, Model, and Active Learning Framework for Small and Rare Wildlife in Aerial Imagery}

\author[1]{\fnm{Bowen} \sur{Zhang}}\email{bowen68@ucsb.edu}

\author[2]{\fnm{Jesse T.} \sur{Boulerice}}\email{boulericejt@si.edu}

\author[1]{\fnm{Charvi} \sur{Mendiratta}}\email{charvi@ucsb.edu}

\author[1]{\fnm{Nikhil} \sur{Kuniyil}}\email{nikhilkuniyil@ucsb.edu}

\author[3]{\fnm{Satish} 
\sur{Kumar}}\email{satishy@stanford.edu}

\author[2,4]{\fnm{Hila} \sur{Shamon}}\email{hshamon@dorisduke.org}

\author[1]{\fnm{B. S.} \sur{Manjunath}}\email{manj@ucsb.edu}

\affil[1]{\orgname{University of California, Santa Barbara}, \orgaddress{\postcode{93106}, \state{California}, \country{United States of America}}}

\affil[2]{\orgname{Smithsonian National Zoo and Conservation Biology Institute}, \orgaddress{\postcode{20008}, \city{Washington, DC}, \country{United States of America}}}

\affil[3]{\orgname{Stanford University}, \orgaddress{\postcode{94305}, \state{California}, \country{United States of America}}}

\affil[4]{\orgname{The Doris Duke Foundation}, \orgaddress{\postcode{10022}, \state{New York}, \country{United States of America}}}

\abstract{
Automated wildlife monitoring from aerial imagery is vital for conservation but remains limited by two persistent challenges: the difficulty of detecting small, rare species and the high cost of large-scale expert annotation. Prairie dogs exemplify this problem—they are ecologically important yet appear tiny, sparsely distributed, and visually indistinct from their surroundings, posing a severe challenge for conventional detection models. To overcome these limitations, 
we present \textbf{RareSpot+}, a detection framework that integrates multi-scale consistency learning, context-aware augmentation, and geospatially guided active learning to address these issues. 
A novel multi-scale consistency loss aligns intermediate feature maps across detection heads, enhancing localization of small ($\sim$ 30 pixels wide) objects without architectural changes, while context-aware augmentation improves robustness by synthesizing hard, ecologically plausible examples. 
A geospatial active learning module exploits domain-specific spatial priors linking prairie dogs and burrows, together with test-time augmentation and a meta-uncertainty model, to reduce redundant labeling.
On a 2\,km$^2$ aerial dataset, RareSpot+ improves detection over the baseline mAP@50 by +35.2\% (absolute +0.13). 
Cross-dataset tests on HerdNet, AED, and several other wildlife benchmarks demonstrate robust detector-level transferability.
The active learning module further boosts prairie dog AP by 14.5\% using an annotation budget of just 1.7\% of the unlabeled tiles.
Beyond detection, RareSpot+ enables spatial ecological analyses such as clustering and co-occurrence, linking vision-based detection with quantitative ecology.

}
%Cross-dataset tests on HerdNet, AED, and several other wildlife benchmarks, demonstrate robust detector transferability. 


\maketitle

\begin{figure*}[t]
    \centering
    \includegraphics[width=\linewidth]{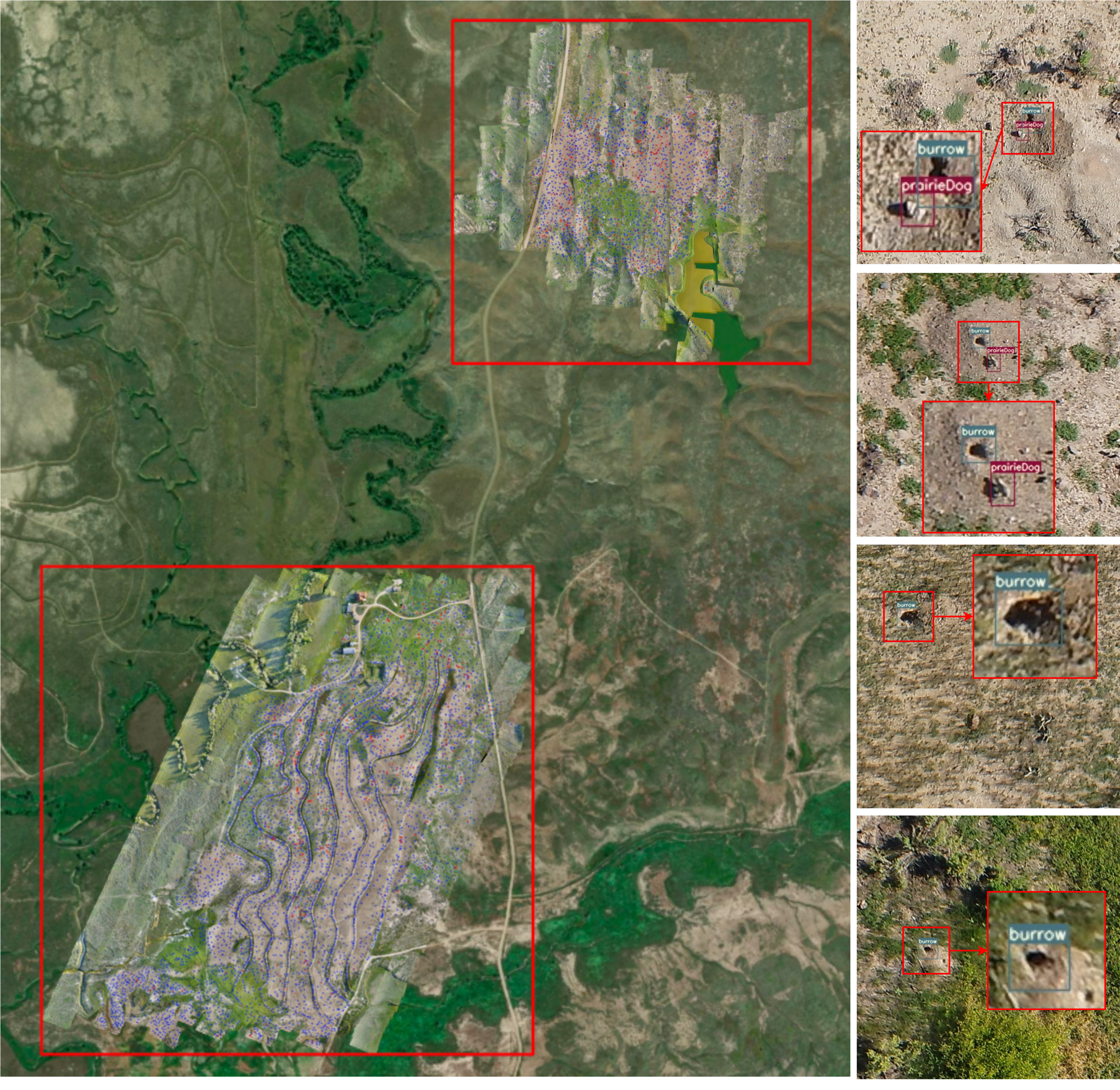}
    \caption{\textbf{Left:} Two large-scale, densely annotated orthomosaics from our drone deployments, separated by 0.8 km and covering a combined area of approximately 2 km\textsuperscript{2}. The red boxes highlight Colony EnrNE (top) and Colony Enr (bottom). \textbf{Right:} Example detections of prairie dogs and burrows. The insets provide zoomed-in views to emphasize the visual difficulty of finding objects from natural background textures such as soil, vegetation, or shadows.}
    \label{fig:overview}
\end{figure*}

\section{Introduction}\label{intro}

Automated wildlife monitoring has become an indispensable tool for ecology and conservation, enabling researchers to estimate species abundance and distribution on broad spatial and temporal scales~\cite{tuia2022perspectives,mcever2023context,kellenberger2018detecting,Delplanque2023ISPRS}. Traditionally, these surveys relied on human observers conducting aerial counts~\cite{Caughley1974JWM,Buckland2001Book}, a process that is labor-intensive, expensive, and prone to observer bias~\cite{Hodgson2018MEE}. The recent adoption of high-resolution drones and aerial imaging platforms has transformed this landscape, allowing efficient data collection over vast regions~\cite{Delplanque2023ISPRS,Gui2024RSODReview}. %However, the sheer volume of data and the difficulty in detecting small, rare species within cluttered natural backgrounds still pose significant challenges for automated analysis.
However, the sheer volume of data, coupled with the difficulty of detecting small, rare species within cluttered natural backgrounds, continues to pose significant challenges for automated analysis and scalable annotation.

Many ecologically important species present two distinct and often conflated challenges for computer vision systems: small visual scale and rare occurrence. We define \emph{small} objects as those occupying only a few pixels in the image, often spanning fewer than $30$ pixels in width, which leads to weak and ambiguous visual cues that are easily confused with soil, vegetation, or shadows under varying illumination and occlusion conditions. In contrast, \emph{rare} objects are defined by their infrequent occurrence across large geographic areas, often accounting for less than $0.1\%$ of the total surveyed area~\cite{Naude_2019_CVPR_Workshops,Huwaterfowl2024}, resulting in extreme class imbalance and high redundancy in manual annotation.
	
Our approach explicitly separates these two phenomena and addresses them with complementary mechanisms. The challenge of small object scale is tackled through representation learning via our \textit{Multi-Scale Consistency Learning} (MSCL) module, which regularizes intermediate feature maps across detection heads to preserve the signal of diminutive targets. The challenge of rare occurrence is addressed through data-efficiency strategies, including \textit{Context-Aware Hard Sample Augmentation} (CA-HSA) and a geospatially guided \textit{Active Learning} (AL) pipeline, which together mitigate class imbalance and substantially reduce redundant human labeling.

%\bowen{Many ecologically important species present a dual challenge for computer vision: they are both small in scale and rare in occurrence. Specifically, we define small objects as those occupying few pixels, often spanning fewer than $30\,pixels$ wide, while rare objects are those that occur infrequently across vast geographic areas, often occupying less than $0.1\%$ of the total surveyed area~\cite{Naude_2019_CVPR_Workshops,Huwaterfowl2024}.\emph{Their weak visual signals are easily confused with soil, vegetation, or shadows, while illumination and occlusion further degrade the detection reliability.} To address these challenges, our work explicitly separates these two phenomena:\begin{itemize}\item \textbf{Small Scale:} Addressed through representation learning via our \textit{Multi-Scale Consistency Learning} (MSCL) module, which regularizes feature maps to preserve the signal of diminutive targets.\item \textbf{Rare Occurrence:} Addressed through data efficiency mechanisms, including \textit{Context-Aware Hard Sample Augmentation} (CA-HSA) and our \textit{Active Learning} (AL) pipeline, which together overcome extreme class imbalance and minimize redundant human annotation.\end{itemize} }

These challenges are exemplified by prairie dogs (\emph{Cynomys} spp.), a keystone genus of burrowing rodents that play a critical role in grassland ecosystems across North America~\cite{Kotliar1999CriticalReview,Miller2000KeystoneResponse,Kotliar2000KeystoneConcept,DavidsonLightfoot2006Ecography,VanNimwegen2008Ecology,Dreelin2025CSP}. Despite their ecological importance, automated detection of prairie dogs in aerial imagery remains an open problem due to the combined effects of limited annotated data, severe class imbalance, and extreme spatial sparsity across large habitats. Moreover, prairie dogs are visually subtle at typical flight altitudes, and their sparse distribution further exacerbates annotation inefficiency and detector bias (Figure~\ref{fig:overview}).

\begin{figure}[t]
    \centering
    \includegraphics[width=\linewidth]{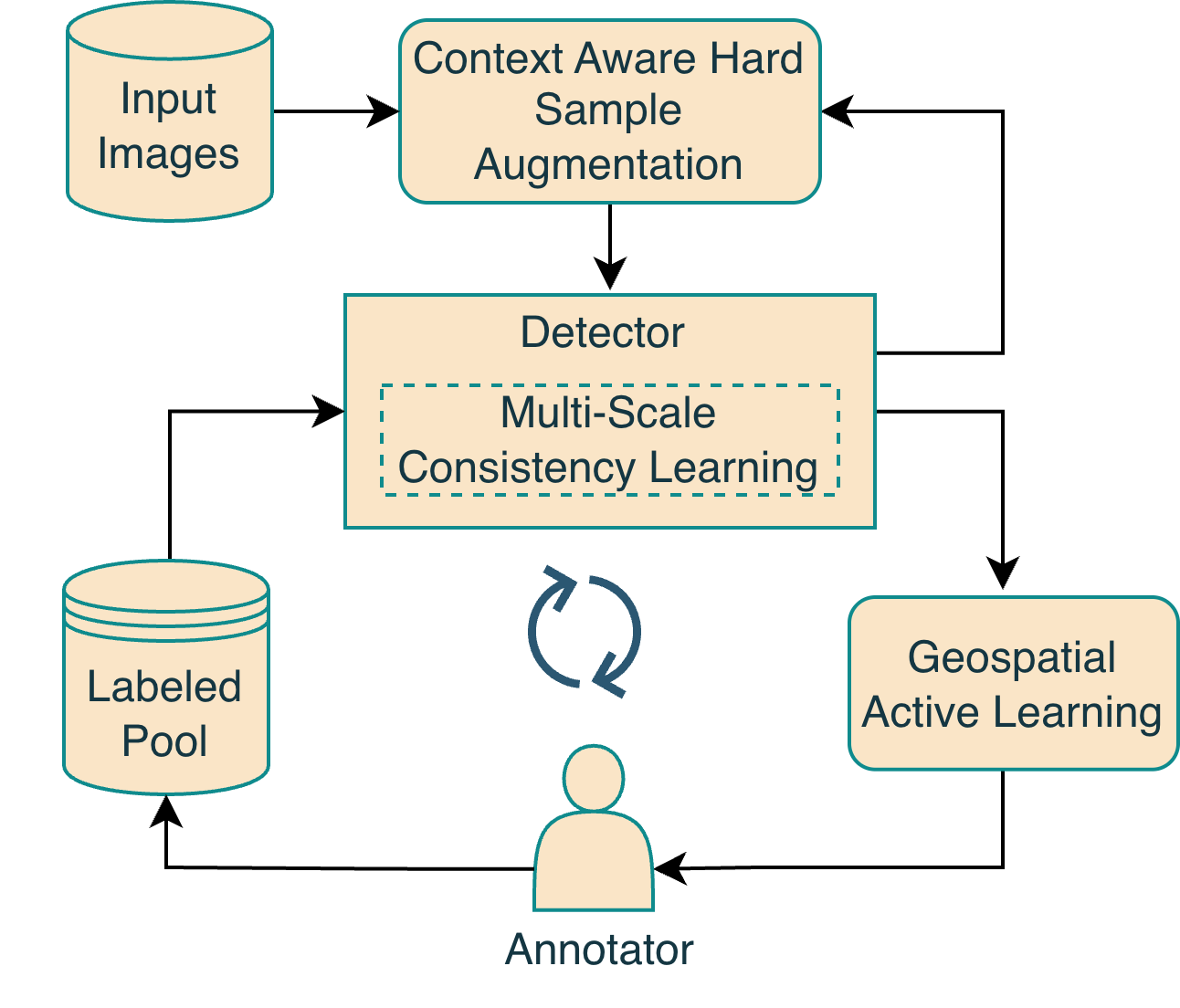} 
    \caption{The RareSpot+ pipeline. This includes Multi-Scale Consistency Learning, Context-Aware Hard Sample Augmentation and Active Learning to overcome data scarcity and rare class imbalance.}
    \label{fig:all_pipeline}
\end{figure}

To address these challenges, we present \textbf{RareSpot+} (Figure~\ref{fig:all_pipeline}), an integrated framework that unifies detection, data augmentation, and active learning strategies for small and rare species in aerial imagery. RareSpot+ refers to the unified framework rather than an individual network architecture, encompassing complementary mechanisms designed to address distinct failure modes in small and rare object detection.

First, a \textit{multi-scale consistency learning} module enforces structured alignment among feature maps from adjacent detection heads, preserving discriminative cues of tiny targets (typically smaller than $32\times32$ pixels, following COCO conventions~\cite{lin2014microsoft}) without altering the underlying network architecture. Second, a \textit{context-aware hard-sample augmentation} process identifies false positives and reinserts them into ecologically plausible backgrounds, improving robustness to habitat-specific confounders. Third, a \textit{geospatial active learning} framework exploits spatial priors—such as the proximity of prairie dogs to burrows—to focus expert labeling effort on the most informative regions, using uncertainty estimates derived from both test-time augmentation and a meta-uncertainty head.

Finally, we introduce the first large-scale, high-resolution aerial dataset of prairie dogs and their burrows. The dataset comprises eight expert-annotated drone surveys covering more than 5~km$^2$ at 2~cm/px resolution, providing a foundation for developing the first automated models for prairie dog and burrow monitoring.

% \paragraph{Positioning of our contribution.} 

%Unlike prior multi-scale fusion approaches such as FPN and PAN that depend on architectural modifications, our method operates entirely at the loss level. The proposed tri-loss regularization aligns adjacent detection heads ($P_3{\leftrightarrow}P_4$, $P_4{\leftrightarrow}P_5$) through MSE, KL, and cosine terms, enhancing small-object representations without adding parameters or inference cost.

%By combining principled feature alignment, context-aware augmentation, and geospatially guided active learning within an openly shared benchmark, RareSpot+ demonstrates how advances in computer vision can directly translate into scalable and reproducible tools for ecological monitoring and conservation.
\paragraph{Positioning of our contribution.}

In contrast to prior multi-scale fusion approaches such as FPN and PAN, which rely on architectural modifications, our approach operates entirely at the loss level. We introduce a tri-loss regularization that aligns adjacent detection heads ($P_3{\leftrightarrow}P_4$ and $P_4{\leftrightarrow}P_5$) using complementary MSE, KL-divergence, and cosine-similarity terms. This formulation strengthens small-object representations during training while introducing no additional parameters at inference-time.

By integrating principled feature-level regularization, context-aware data augmentation, and geospatially guided active learning within an openly released benchmark, RareSpot+ illustrates how advances in computer vision can translate into scalable, reproducible tools for ecological monitoring and conservation.

\section{Related Work}\label{sec:related_work}

Prairie dog colonies and their extensive burrow systems play a critical ecological role by contributing to prey availability, landscape heterogeneity, nutrient cycling, and subterranean habitats that sustain diverse dependent species, including insects \cite{DavidsonLightfoot2006Ecography}, birds \cite{SmithLomolino2004Oecologia,Dinsmore2005JWM,Dreelin2025CSP}, reptiles \cite{KretzerCully2001SWNat}, and mammals \cite{Hoogland1995Book,Kotliar2000KeystoneConcept,DavidsonLightfoot2006Ecography,Stapp2007Jmamm}, notably the endangered black-footed ferret (\textit{Mustela nigripes}) \cite{Kotliar1999CriticalReview,Eads2011Jmamm}. Although previous UAV-based efforts have mapped the distribution of prairie dog burrows \cite{Delparte,Kearney}, these studies were limited to static burrow identification and did not address direct detection of prairie dogs themselves. 
Our work fills this gap by introducing the first framework and benchmark dataset for joint detection of prairie dogs and their burrows, enabling automated monitoring of colony activity. Furthermore, the natural co-occurrence of prairie dogs and burrows provides a principled spatial cue for uncertainty estimation in active learning and for analyzing spatiotemporal variations in colony dynamics.

%Our work fills this gap by introducing the first end-to-end framework and benchmark dataset for simultaneous detection of prairie dogs and their burrows, enabling automated monitoring of colony activity. Furthermore, the co-occurrence of prairie dogs and burrows provides a natural cue for uncertainty estimation in active learning and for assessing spatiotemporal variations in colony dynamics.

\subsection{Small-Object Detection in Aerial and Ecological Imagery}
Detecting small objects remains a persistent challenge in aerial and ecological imagery, where target instances occupy only a few dozen pixels and are often indistinguishable from background clutter. Conventional detection frameworks such as Faster R-CNN~\cite{faster_rcnn} and YOLO~\cite{redmon2018yolov3,bochkovskiy2020yolov4} have demonstrated remarkable success on generic benchmarks, but degrade significantly when object size falls below 32 pixels~\cite{zhu2021tph,wang2022yolov7}. Several methods have attempted to improve feature resolution through multi-scale feature fusion~\cite{lin2017feature,tan2020efficientdet,wang2022yolov7} or by incorporating transformer-based global reasoning~\cite{carion2020detr,swin}. Nevertheless, these architectures often struggle to retain fine-grained cues critical for small-object localization, particularly when trained with limited data.

Building on these general challenges, ecological detection tasks pose additional difficulties due to cluttered natural backgrounds, low inter-class variability, and strong dependence on environmental context. Prior studies have focused primarily on adapting generic detectors to specific taxa, such as elephants~ \cite{Naude_2019_CVPR_Workshops}, waterfowl~\cite{Huwaterfowl2024}, and seals~\cite{laradji2022sealnet}, or on leveraging multi-sensor cues such as thermal and hyperspectral imagery~\cite{chabot2018canada,Gui2024RSODReview}. While these approaches have improved detection accuracy for specific species, they rely heavily on extensive annotation or strong priors about object size and appearance. In contrast, our work introduces a \textit{multi-scale consistency loss} that regularizes feature maps across adjacent detection heads, improving the representation of tiny objects without modifying the backbone architecture or adding new parameters during inference, requiring only a minimal increase in parameters during training.

\paragraph{Relation to prior multi-scale and consistency learning}
While our approach regularizes feature hierarchies through the loss function, existing multi-scale fusion approaches such as FPN, PAN, and transformer-based detectors propagate information across scales through architectural changes \cite{zhang2025sodetr, du2024crosslayer}, often increasing model complexity to stabilize representations of small objects. 
In contrast, our formulation operates at the \emph{loss level}, introducing a lightweight tri-loss that regularizes existing detection heads.
While prior representation-stability methods~\cite{huang2022small,huynh2024one,simeoni2025dinov3} focus on scale-invariant features or patch-level consistency, we explicitly enforce \emph{cross-head coherence} by aligning adjacent levels ($P_3{\leftrightarrow}P_4$, $P_4{\leftrightarrow}P_5$). 
This adjacency-based design preserves fine-scale cues critical for tiny objects while stabilizing higher-level semantics, as illustrated in Fig.~\ref{fig:feature_map}.

\begin{table*}[t]
    \centering
  %  \caption{Summary of prairie dog habitat datasets. The table lists the number of annotated and total tiles, prairie dogs (PD), burrows, and orthomosaic dimensions across training, validation, and test splits. \textbf{AL} denotes datasets used for active learning, and \bowen{\textbf{Partial} refers to datasets with ongoing annotations, as this work also serves as a dataset and benchmark paper; these rows are not used in the experiments presented in this study but are intended to support future active learning or weakly supervised learning research.} The prefixes \textit{EnrNE} and \textit{Enr} in the dataset names refer the two colonies.}
  \caption{Summary of prairie dog habitat datasets. The table reports the number of annotated and total tiles, prairie dogs (PD), burrows, and orthomosaic dimensions across training, validation, and test splits. \textbf{AL} denotes datasets used for active learning. \textbf{Partial} indicates datasets with ongoing annotations that are released as part of the benchmark but are not used in the experiments reported in this paper, and are intended to support future active learning and weakly supervised learning research. The prefixes \textit{EnrNE} and \textit{Enr} in the dataset names refer to the two colonies.}

    \label{tab:data_split}
    \setlength{\tabcolsep}{4pt}
    \begin{tabularx}{\linewidth}{@{} l l *{5}{>{\centering\arraybackslash}X} @{}} 
        \toprule
        \textbf{{Dataset}} & \textbf{{Data Type}} & \textbf{{\# Annot. Tiles}} & \textbf{{\# Total Tiles}} & \textbf{{\# PD}} & \textbf{{\# Burrows}} & \textbf{{Orthomosaic Size (px)}} \\ 
        \midrule
        \rowcolors{2}{gray!15}{white}
        EnrNE\_fly\_1 & Train       & 7{,}858 & 33{,}294 & 807 & 3{,}913 & 95K$\times$91K \\
        EnrNE\_fly\_2 & Validation  & 9{,}293 & 27{,}888 & 569 & 4{,}435 & 85K$\times$84K \\
        Enr\_fly\_3   & Test        & 7{,}616 & 55{,}224 & 609 & 4{,}435 & 120K$\times$119K \\
        Enr\_fly\_4   & AL          & 3{,}534       & 58{,}560 & 771 & 3{,}572 & 122K$\times$124K \\
        EnrNE\_fly\_5 & Partial     & 960       & 16{,}240 & 161 & 968     & 57K$\times$73K \\
        EnrNE\_fly\_6 & Partial     & 1{,}434       & 27{,}547 & 91  & 1{,}491 & 86K$\times$83K \\
        EnrNE\_fly\_7 & Partial     & 1{,}266       & 28{,}728 & 127 & 1{,}331 & 87K$\times$85K \\
        Enr\_fly\_8   & Partial     & 2{,}072       & 14{,}762 & 101 & 2{,}590 & 61K$\times$62K \\
        \bottomrule
    \end{tabularx}
\end{table*}

\subsection{Context-Aware Data Augmentation}
Data augmentation is a well-established strategy for improving robustness and diversity in training data~\cite{shorten2019survey,zhang2018mixup,yun2019cutmix}. Recent work has explored techniques such as synthetic copy--paste~\cite{ghiasi2021copypaste}, style transfer~\cite{georgakis2017synthesizing}, and adversarial augmentation~\cite{volpi2018generalizing} to mitigate domain imbalance. However, these generic augmentation strategies rarely account for the ecological context in which small objects appear. In wildlife monitoring scenarios, spatial and textural cues—including vegetation density, soil coloration, and shadow patterns—strongly influence false detections~\cite{kellenberger2018detecting,tuia2022perspectives}.

Our \textit{context-aware hard-sample augmentation} strategy generates hard yet ecologically realistic training examples that improve model generalization. Unlike prior copy--paste approaches, object placement is guided by learned habitat context rather than random spatial overlays of ground-truth instances, thereby better reflecting the environmental variability encountered in aerial wildlife surveys.

\subsection{Active Learning for Efficient Annotation}
%Active learning (AL) offers a principled approach to reduce labeling effort by iteratively selecting the most informative samples for annotation. Common selection criteria include model uncertainty~\cite{li2024survey}, diversity~\cite{Yamani_2024_WACV}, or hybrid formulations that balance both~\cite{ash2019deep}. In computer vision, AL has been applied to object detection~\cite{roy2018deep,kasarla2019region,siddiqui2020viewal}, segmentation~\cite{mackowiak2018cereals}, and remote sensing~\cite{tuia2009active}. Teacher–student frameworks have combined AL with semi-supervised detection to refine pseudo-labels \cite{Zhang_2024}, and Few-shot and low-shot methods have also been extended with AL to handle rare species and long-tailed class distributions \cite{rs17071155, GuirguisEWKM0B22}, which are particularly relevant in ecological monitoring. Yet, most existing AL strategies ignore the strong spatial dependencies present in ecological imagery, where animal occurrences are correlated with habitat structures or co-located features such as nests, burrows, or vegetation patches.

Active learning (AL) provides a principled framework for reducing labeling effort by iteratively selecting the most informative samples for annotation. Common selection criteria include model uncertainty~\cite{li2024survey}, sample diversity~\cite{Yamani_2024_WACV}, and hybrid strategies that balance both~\cite{ash2019deep}. In computer vision, AL has been applied to object detection~\cite{roy2018deep,kasarla2019region,siddiqui2020viewal}, segmentation~\cite{mackowiak2018cereals}, and remote sensing~\cite{tuia2009active}. Teacher--student frameworks have further combined AL with semi-supervised detection to refine pseudo-labels~\cite{Zhang_2024}, while few-shot and low-shot methods augmented with AL have been explored to address rare species and long-tailed class distributions~\cite{rs17071155,GuirguisEWKM0B22}, which are particularly relevant in ecological monitoring. Nevertheless, most existing AL strategies treat samples as independent and ignore the strong spatial dependencies characteristic of ecological imagery, where animal occurrences are often correlated with habitat structures or co-located features such as nests, burrows, or vegetation patches.

We address this limitation through a \textit{geospatial active learning} framework that incorporates species--habitat spatial priors to prune the candidate annotation pool prior to uncertainty-based ranking. In our case study, prairie dogs exhibit strong spatial association with burrow locations; leveraging this relationship enables us to prioritize labeling of tiles most likely to contain animals. We further integrate test-time augmentation and a meta-uncertainty model to refine sample selection. Together, these components substantially reduce annotation cost without sacrificing detection accuracy.

In summary, prior advances in small-object detection, data augmentation, and active learning have largely evolved independently of ecological applications. \textbf{RareSpot+} unifies these directions within a single domain-aware framework, providing a reproducible benchmark for large-scale aerial wildlife monitoring and establishing a foundation for translating computer vision innovations into ecological analysis.

\section{Prairie Dog Benchmark Dataset}\label{sec:dataset}
Prairie dogs (\textit{Cynomys spp.}) are keystone species that structure North American grassland ecosystems by aerating soil, altering vegetation, and serving as prey for numerous predators. Accurate monitoring of colony distribution and density is therefore central to assessing ecosystem health. Traditional ground surveys are labor-intensive and spatially limited, motivating the use of high-resolution aerial imagery for scalable and repeatable monitoring.

We constructed our benchmark dataset from eight aerial surveys of two prairie dog colonies, covering approximately $5~\text{km}^2$ of mixed-grass prairie (Table~\ref{tab:data_split}). Flights were conducted during the summer months on multiple dates and times under stable illumination using a fixed-wing drone (SenseFly eBee X) equipped with a 24~MP RGB camera (SenseFly Aeria X). All surveys were flown at an above-ground altitude of approximately 100~m, yielding an average ground sampling distance (GSD) of 2~cm/px, with 75\% lateral and 70\% longitudinal overlap between adjacent flight lines. Orthomosaics were generated using Pix4D software, producing georeferenced mosaics with centimeter-level spatial accuracy.

\subsection{Image Tiling and Annotation Workflow}
Each orthomosaic was subdivided into fixed-size tiles (512$\times$512\,px) to facilitate efficient annotation and model training. Annotation was performed using the BisQue platform~\cite{bisque} and cross-validated by multiple experts. Two object classes were labeled: (i)~\textit{prairie dogs}, identified by characteristic body shape and shadow patterns, and (ii)~\textit{burrows}, visible as circular openings with surrounding soil mounds.

To ensure annotation consistency, a two-stage quality-control process was employed. First, nine trained annotators independently labeled and cross-checked the data. This was followed by senior ecologist verification across all annotated tiles. Disagreements were jointly reviewed to refine labeling guidelines and resolve ambiguous cases.

%Each orthomosaic was subdivided into fixed-size tiles (512$\times$512\,px) for efficient annotation and model training. Annotation was performed in the BisQue platform~\cite{bisque} and cross-validated by multiple experts. Two object classes were labeled: (i)~\textit{prairie dogs}, identified by characteristic body shape and shadow pattern, and (ii)~\textit{burrows}, visible as circular openings with soil mounds. To ensure consistency, a two-stage quality-control process was used: first, nine trained annotators performed and cross-checked the initial labeling, followed by an senior ecologist verification on all data patches. Disagreements were reviewed jointly to refine the labeling guidelines.

\subsection{Dataset Statistics and Splits}
%The final dataset contains 34{,}033 annotated tiles with a total of 3{,}236 prairie dog and 22{,}735  burrow instances, covering a total area of approximately 5~km\textsuperscript{2}. Prairie dog bounding box sizes in these images range from 15–80\, pixels in length (mean $\sim$30\,px), representing a challenging small-object regime. 
%\bsm{so these are about 600 to 900 pixels in area? it is not clear when we say sub-30 pixels whether we are referring to area or length. 600-900 pixels is quite big.} 
The final dataset comprises 34{,}033 annotated tiles containing a total of 3{,}236 prairie dog instances and 22{,}735 burrow instances, covering approximately 5~km\textsuperscript{2} of surveyed area. Prairie dog bounding boxes range from 15 to 80 pixels in \emph{maximum side length}, with a mean of approximately 30 pixels, placing this dataset firmly within a challenging small-object detection regime.

To evaluate cross-site generalization, the data are partitioned \emph{geographically} into two non-overlapping prairie dog colonies: Colony EnrNE (Enrico North East) and Colony Enr (Enrico), separated by approximately 0.8~km and differing in vegetation density and soil coloration (Figure~\ref{fig:overview}). All training and validation data are drawn exclusively from Colony EnrNE, while Colony Enr is reserved for testing, ensuring strict spatial separation between training and evaluation regions. Within each colony, orthomosaics are tiled into $512\times512$ pixel patches with 30\% overlap. All tiles containing at least one annotation are included, along with an equal number of randomly sampled background tiles without objects, to construct the detection training set. A detailed statistical breakdown of all eight datasets is provided in Table~\ref{tab:data_split}. Datasets labeled as \textbf{Partial} correspond to surveys with ongoing annotations; these data are not used in the experiments reported here but are released to support future research in active learning and weakly supervised detection.

%To evaluate cross-site generalization, the data are partitioned \textit{geographically} into two non-overlapping colonies: Colony EnrNE (Enrico North East) and Colony Enr (Enrico). They are separated by 0.8 km and differing in vegetation density and soil coloration Figure~\ref{fig:overview}.  All training and validation images are drawn from Colony EnrNE, while Colony Enr is reserved for testing, ensuring no spatial overlap between training and testing regions. Within each colony, images are tiled into $512 \times 512$ patches with a 30\% overlap. \bowen{All tiles containing annotations are included, along with an equal number of randomly sampled tiles without objects, to construct the detection training pipeline.} A detailed statistical breakdown for each of the eight datasets is provided in Table~\ref{tab:data_split}. \bowen{The term \textbf{Partial} refers to datasets with ongoing annotations. These rows are excluded from our current experiments but are provided to support future research in active or weakly supervised learning.}

\begin{figure}[t]
    \centering
    \includegraphics[width=\linewidth]{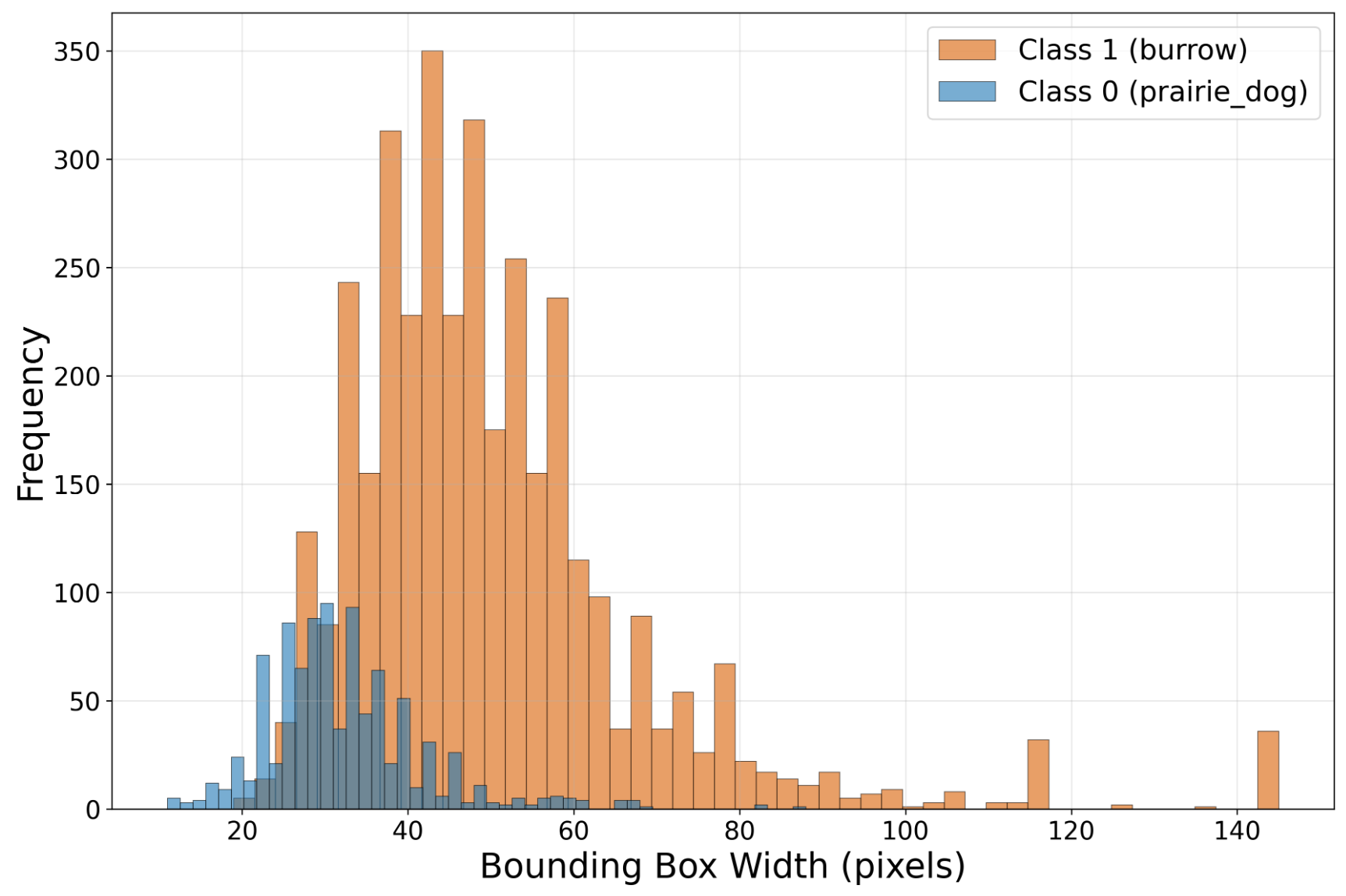} 
    \caption{Dataset statistics of the training dataset.}
    \label{fig:dataset_stats}
\end{figure}

% \bsm{(can we add such a figure? would be good to see this distribution if possible. otherwise delete this.)} 
%Figure~\ref{fig:dataset_stats} shows the frequency distribution of the widths of the bounding box for prairie dogs and burrows, illustrating the range of object sizes and highlighting the challenge of detecting small objects and imbalanced classes. 
% The two colonies also differ in burrow density and vegetation coverage, providing natural variability for evaluating spatial generalization. 
%Compared with existing aerial wildlife datasets such as AED~\cite{Naude_2019_CVPR_Workshops}, Waterfowl~\cite{Huwaterfowl2024}, and WAID~\cite{kumar2024wildlifemapper}, our benchmark provides higher spatial resolution and explicit coupling between animal and habitat features, enabling rigorous evaluation of spatial generalization.
Figure~\ref{fig:dataset_stats} shows the frequency distributions of bounding box widths for prairie dogs and burrows, illustrating the wide range of object sizes and highlighting both the challenge of detecting small objects and the severe class imbalance present in the data. Compared with existing aerial wildlife datasets such as HerdNet~\cite{Delplanque2023ISPRS}, Eikelboom et al.~\cite{eikelboom2019improving},  AED~\cite{Naude_2019_CVPR_Workshops}, Waterfowl~\cite{Huwaterfowl2024}, and WAID~\cite{kumar2024wildlifemapper}, our benchmark offers higher spatial resolution and an explicit coupling between animal and habitat features, enabling systematic evaluation of detector performance under realistic spatial variability.

\begin{figure*}[t]
    \centering
    \includegraphics[width=0.95\linewidth]{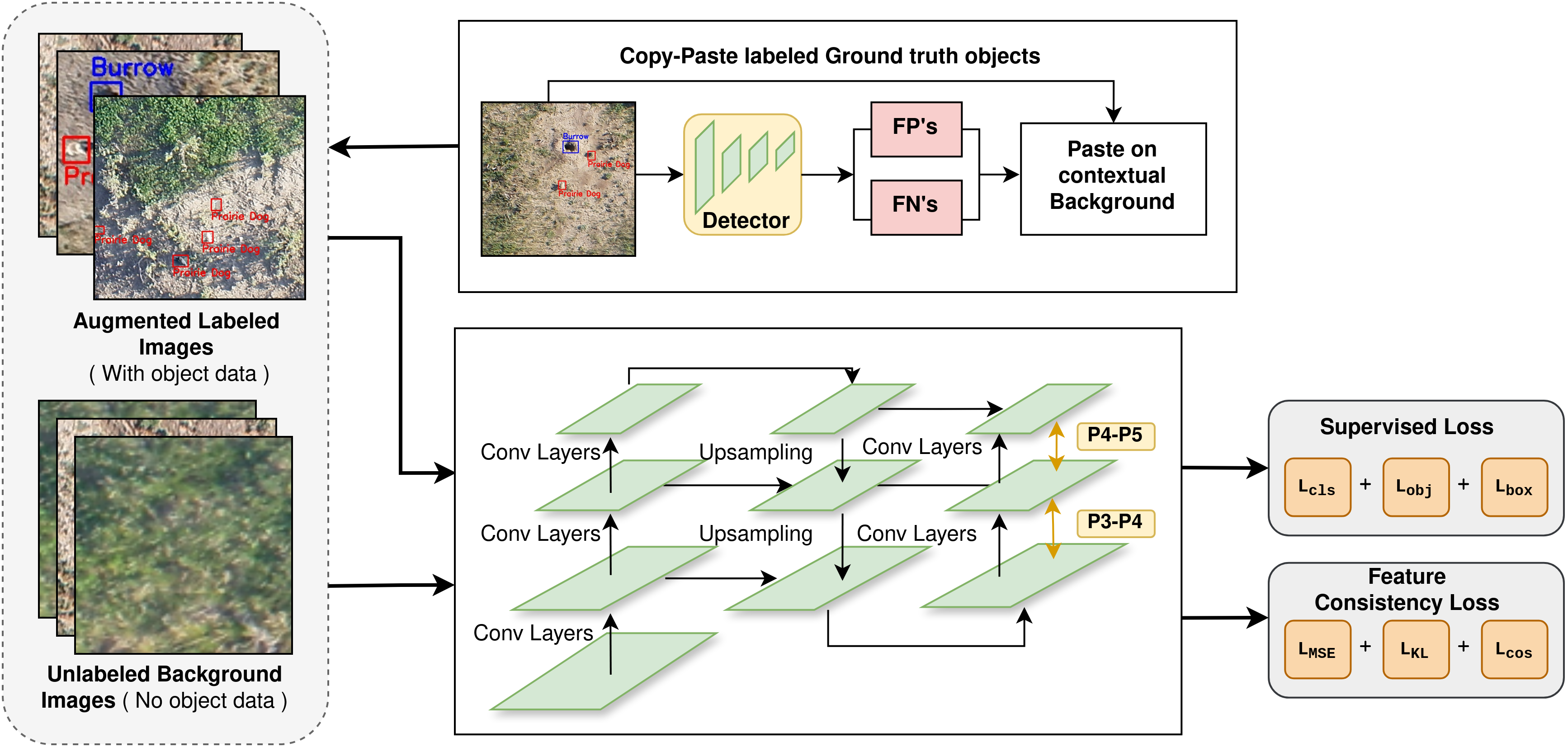} 
    \caption{Multi-Scale Consistency Learning applied across feature layers and Context-Aware Hard Sample Augmentation.}
    \label{fig:detection}
\end{figure*}
\subsection{Ecological Utility and Public Release}
Beyond serving as a computer vision benchmark, the dataset supports downstream ecological analyses such as colony density estimation, spatial clustering, and co-occurrence analysis between prairie dogs and burrows (Section~3 in Supplementary Materials). These dual-use capabilities make the dataset valuable to both the machine learning and ecological research communities.

The dataset will be publicly released at the time of publication and will include all imagery, annotations, and standardized data splits used in this study. To protect sensitive habitats, geolocation metadata will be anonymized prior to release. All aerial surveys were conducted under approved institutional wildlife-monitoring permits and in compliance with local regulations, ensuring minimal disturbance to prairie dog colonies during data acquisition. The data and accompanying code will be made available through Bisque UCSB~\cite{rarespot_dataset}
% \href{https://bisque2.ece.ucsb.edu/client_service/view?resource=https://bisque2.ece.ucsb.edu/data_service/00-B2mLaCeFPkCgZF4LND9uEd}{Bisque UCSB}.

% \url{https://tinyurl.com/BisqueUCSB}

% \url{https://github.com/eebowen/RareSpot}.

% \bsm{We should put the data on BisQue. Why is it on Github??}

%Beyond serving as a computer-vision benchmark, the dataset enables ecological analyses such as colony density estimation, spatial clustering, and co-occurrence between prairie dogs and burrows (Section~\ref{sec:geo_analysis}). These dual-use capabilities make it valuable to both machine-learning and ecological communities. The dataset will be publicly released at the time of publication, including all images, annotations, and standardized data splits. Geolocation metadata will be anonymized to protect sensitive habitats. All flights were conducted under institutional wildlife-monitoring permits in compliance with local regulations, ensuring minimal disturbance to the colonies during data acquisition. The data and code will be available at the project's repository: \url{https://github.com/eebowen/RareSpot}

\section{The RareSpot+ Detection Method}
\label{sec:methods}

Robust recognition and detection often rely on multi-scale feature processing, which enables models to handle objects spanning a wide range of sizes~\cite{huang2022small,huynh2024one}. Recent work such as DINOv3~\cite{simeoni2025dinov3} highlights this principle through \textit{Gram anchoring}, which stabilizes patch-level representations across scales. This idea is closely related to the fine-grained feature consistency explored in earlier versions of this work~\cite{zhang2025rarespotspottingsmallrare}, where preserving high-frequency spatial detail within feature maps was shown to be critical for improving detection and segmentation accuracy. Figure~\ref{fig:detection} provides an overview of the RareSpot+ detection method, which integrates a convolutional backbone with a multi-scale detection head and a context-aware augmentation strategy to enhance recognition of small and rare objects in aerial imagery.

%Robust recognition and detection often rely on multiscale feature processing, which allows models to handle objects of varying sizes~\cite{huang2022small,huynh2024one}. Recent work such as DINOv3~\cite{simeoni2025dinov3} highlights this through \textit{Gram anchoring}, which stabilizes patch-level representations across scales. This principle aligns closely with the fine-grained feature consistency introduced in the prior version of this work~\cite{zhang2025rarespotspottingsmallrare}, both aiming to preserve high-frequency spatial detail within feature maps to improve detection and segmentation accuracy. Figure~\ref{fig:detection} illustrates the overall RareSpot framework, which integrates a convolutional backbone with a multiscale detection head and a context-aware augmentation strategy designed to enhance rare and small-object recognition in aerial imagery.

\subsection{Multi-Scale Consistency Learning} 
We introduce \textit{multi-scale consistency learning}, a lightweight tri-loss alignment mechanism that enforces structured correspondence across adjacent feature scales using Mean Squared Error (MSE), Kullback--Leibler (KL) divergence, and cosine similarity. By explicitly regularizing feature representations across detection heads, this approach is particularly effective for the challenging task of tiny object detection.

%In this section, we introduce \textit{multi-scale consistency learning}, a lightweight tri-loss alignment mechanism that enforces structured correspondence across scales using Mean Squared Error (MSE), Kullback--Leibler (KL) Divergence, and Cosine Similarity. This loss is applied only during training, adding no overhead during inference, and is particularly effective for the challenging task of tiny-object detection.

\subsection{Feature Alignment Across Scales}
To enforce cross-scale consistency, we explicitly align feature maps extracted from the YOLOv5 FPN--PAN hierarchy. Let $P_3 \in \mathbb{R}^{C \times H \times W}, 
P_4 \in \mathbb{R}^{C \times \frac{H}{2} \times \frac{W}{2}}, 
P_5 \in \mathbb{R}^{C \times \frac{H}{4} \times \frac{W}{4}},$
denote the feature maps at three successive pyramid levels, where \(C\) is the number of channels and \((H, W)\) are the spatial dimensions of \(P_3\). To enable direct comparison across scales, we upsample \(P_4\) and \(P_5\) to the highest spatial resolution, yielding
$\tilde{P}_4, \tilde{P}_5 \in \mathbb{R}^{C \times H \times W}.$

\noindent\paragraph{Mean Squared Error (MSE):}
The MSE term promotes similarity in raw activation magnitudes between adjacent feature scales, encouraging point-wise alignment across the feature pyramid:
%MSE promotes similarity in raw activation magnitudes between adjacent scales:
{\footnotesize
\begin{equation}
\begin{aligned}
\mathcal{L}_{\mathrm{MSE}} =
\frac{1}{HW} \sum_{i=1}^H \sum_{j=1}^W
\Bigl\| P_3^{(i,j)} - \tilde{P}_4^{(i,j)} \Bigr\|^2 \\
+ \Bigl\| \tilde{P}_4^{(i,j)} - \tilde{P}_5^{(i,j)} \Bigr\|^2.
\end{aligned}
\end{equation}
}
This loss enforces consistent activation strength for potential targets across adjacent pyramid levels, helping to prevent tiny objects from being suppressed or lost as features propagate between scales.

%\bowen{Encourages point-wise magnitude alignment, ensuring that the activation strength for a potential target remains consistent across adjacent scales. This prevents the detector from losing a tiny object when it transitions between feature pyramid levels.}

\noindent\paragraph{Kullback--Leibler (KL) Divergence:}
To complement the MSE term, which enforces numerical alignment, the KL divergence encourages semantic coherence across scales by penalizing cases in which adjacent feature levels produce qualitatively different spatial distributions for the same image region. This encourages neighboring detection heads to focus on the same semantic center of a small animal target. Treating feature vectors as probability distributions via softmax normalization, we define
%To complement MSE, which enforces numerical alignment, KL divergence promotes semantic coherence across scales. \bowen{It penalizes instances where two adjacent scales produce qualitatively different spatial distributions for the same region, ensuring that both heads focus on the same semantic center of the small animal target.} Treating feature vectors as probability distributions via softmax normalization, we define
{
\begin{equation}
\begin{aligned}
\mathcal{L}_{\mathrm{KL}} =
\frac{1}{HW} \sum_{i=1}^H \sum_{j=1}^W
\mathrm{KL}\bigl(S[P_3]^{(i,j)},\, S[\tilde{P}_4]^{(i,j)}\bigr) \\
+ \mathrm{KL}\bigl(S[\tilde{P}_4]^{(i,j)},\, S[\tilde{P}_5]^{(i,j)}\bigr),
\end{aligned}
\end{equation}
}
\noindent where $\mathrm{KL}(\cdot,\cdot)$ denotes the Kullback--Leibler divergence and $S[\cdot]$ represents softmax normalization.
%\noindent where $\mathrm{KL}(\cdot,\cdot)$ denotes KL divergence and $S[\cdot]$ represents softmax normalization.

\noindent\paragraph{Cosine Similarity:}
% Finally, cosine similarity preserves angular relationships between feature vectors, maintaining orientation consistency even when small-object features weaken:
%\bowen{Finally, cosine similarity focuses on the directional orientation of feature vectors. By enforcing angular consistency, we ensure the underlying structural representation of the target is preserved:}
Finally, cosine similarity focuses on the directional orientation of feature vectors across scales. By enforcing angular consistency, this term helps preserve the underlying structural representation of small targets, even when absolute activation magnitudes weaken.

{\footnotesize
\begin{equation}
\begin{aligned}
\mathcal{L}_{\mathrm{cos}} =
\frac{1}{HW} \sum_{i=1}^H \sum_{j=1}^W
\Bigl[1 - \cos\bigl(P_3^{(i,j)},\, \tilde{P}_4^{(i,j)}\bigr)\Bigr] \\
+ \Bigl[1 - \cos\bigl(P_4^{(i,j)},\, \tilde{P}_5^{(i,j)}\bigr)\Bigr].
\end{aligned}
\end{equation}
}

\subsection{Final Multi-Scale Consistency Loss}
We combine the three components into a unified objective:
{
\begin{equation}
\mathcal{L}_{\mathrm{consistency}} =
\alpha \,\mathcal{L}_{\mathrm{MSE}}
+ \beta \,\mathcal{L}_{\mathrm{KL}}
+ \gamma \,\mathcal{L}_{\mathrm{cos}}
\end{equation}
}
\noindent 
Here, \(\mathcal{L}_{\mathrm{MSE}}\) minimizes numerical discrepancies in activation magnitude, \(\mathcal{L}_{\mathrm{KL}}\) aligns semantic distributions across scales, and \(\mathcal{L}_{\mathrm{cos}}\) preserves angular coherence in feature orientation. Together, these complementary terms encourage stable and consistent representations of small objects across adjacent pyramid levels, improving detection robustness in cluttered aerial imagery.

In our fixed-altitude aerial survey setting, prairie dog targets consistently appear at small spatial scales. The proposed consistency loss explicitly leverages this property by reinforcing cross-scale coherence in a manner tuned for small-object detection. Each alignment term is assigned an independent weighting coefficient, allowing fine-grained control over its contribution and selective emphasis across pyramid levels. This design preserves the spatial precision of lower-level features (\(P_3\)) and the semantic abstraction of higher-level features (\(P_5\)), while improving consistency between them. The consistency loss is applied jointly with the standard detection objectives—bounding box regression, objectness, and classification losses—using tunable coefficients to ensure compatibility with the overall detection framework.

\paragraph{Design rationale and exact pairing}
Consistency alignment is restricted to \emph{adjacent feature pairs} ($P_3{\leftrightarrow}P_4$ and $P_4{\leftrightarrow}P_5$), where complementary spatial and semantic information is most strongly correlated. The tri-loss weighting coefficients are set to $(\alpha,\beta,\gamma) = (10.0,\,1.0,\,1.0)$ and are applied uniformly to both adjacent pairs. Furthermore, a detailed loss ablation and sensitivity analysis are provided in Table 1 of the Supplementary Materials. The consistency objective is used \emph{only during training} and introduces no additional parameters at inference time, thereby preserving the computational efficiency of the baseline detector.

\begin{figure*}[t]
    \centering
    \includegraphics[width=0.9\linewidth]{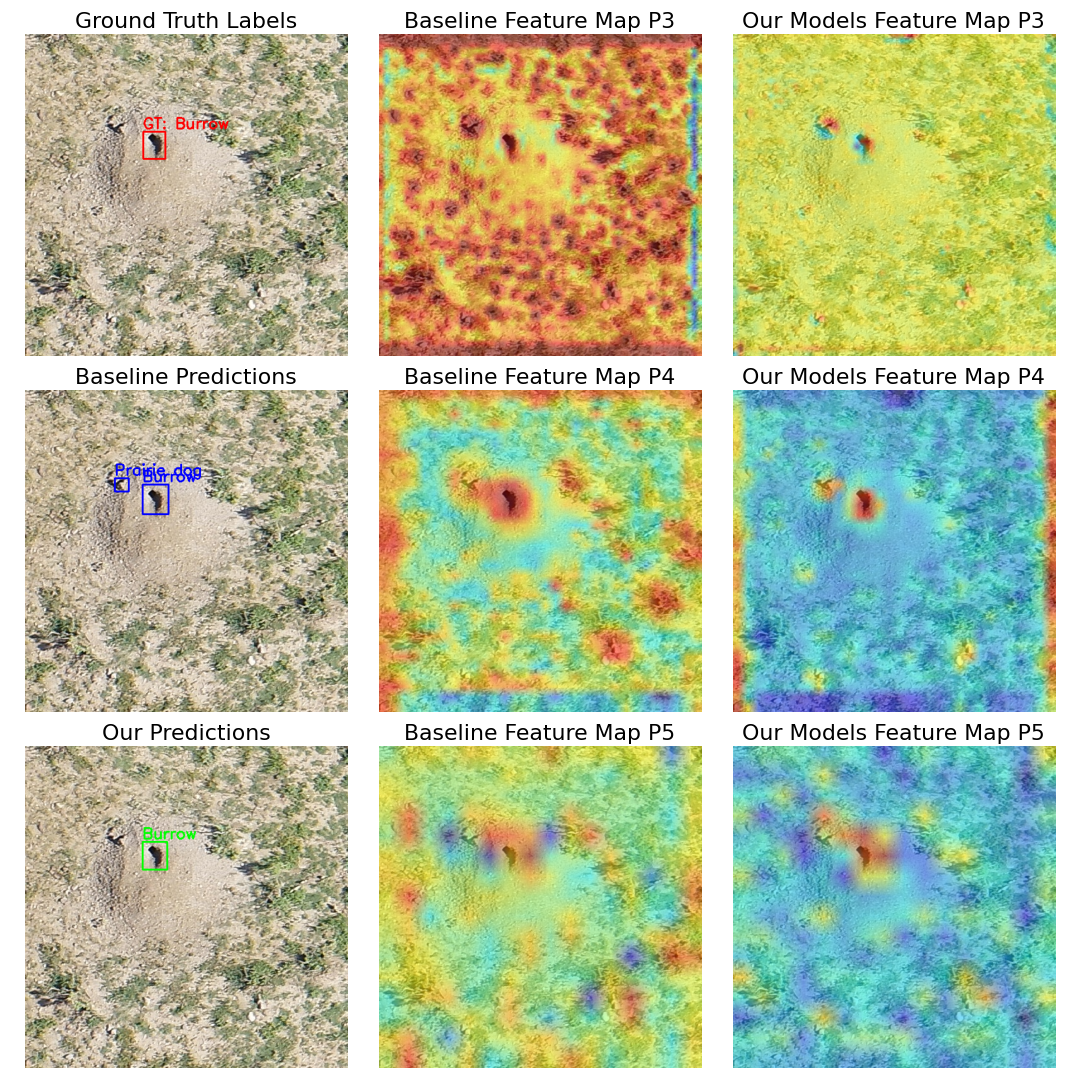}
    \caption{Comparison of the RareSpot+ model (ours) with a baseline. The first column shows the ground truth and the resulting predictions at 0.2 confidence threshold. The second and third columns present the corresponding feature maps from both models across different feature levels (P3, P4, and P5). Multi-Scale Consistency Learning in our model produces more focused and localized feature activations, improving object localization compared to the baseline's noisy feature maps.}
    \label{fig:feature_map}
\end{figure*}

\subsection{Context-Aware Hard Sample Augmentation for Rare and Small Object Detection}
\label{sec:aug}
Prairie dogs present an extreme small-object detection challenge in drone imagery: they constitute only 0.81\% of the annotated instances in our dataset and exhibit weak visual signals compared with objects in large-scale benchmarks such as COCO. To address this, we propose a \textit{context-aware hard sample augmentation} (CA-HSA) strategy that leverages model errors and ecological context to improve robustness and generalization.

Our approach begins by mining hard examples from model predictions, specifically false positives (FPs) and false negatives (FNs), which typically correspond to ambiguous visual patterns such as shadows, rocks, or vegetation textures. These samples are combined with true labeled instances to form a pool of challenging patches. Each patch is then embedded into a \emph{semantically consistent} background to generate realistic yet difficult training examples.

Backgrounds are constructed by sampling empty (background-only) tiles from the training set and applying HSV color-based segmentation to separate habitat regions into dirt/mud and grass. This process yields a pixel-wise \emph{context map} \(c\), where \(c(i,j) \in \{\text{dirt}, \text{grass}\}\) denotes the local habitat type.

Patches drawn from the sets \(P_{\text{Labeled}}, P_{\text{FP}},\) and \(P_{\text{FN}}\) are placed into contextually matched regions—90\% in dirt and 10\% in grass—reflecting their empirical distribution in the dataset. Each patch undergoes random transformations parameterized by
\( \theta = \{\theta_{\text{scale}}, \theta_{\text{rot}}, \theta_{\text{illum}}\} \):

\begin{itemize}
    \item $\theta_{\text{scale}}$: random scaling in $[0.9, 1.1]$
    \item $\theta_{\text{rot}}$: random in-plane rotation in  $[-90^\circ, +90^\circ]$
    \item $\theta_{\text{illum}}$: random brightness and contrast adjustment in HSV space 
\end{itemize}

Augmented patches are blended into the background using Poisson image blending via the \texttt{SeamlessClone} module in OpenCV~\cite{opencv_library}, ensuring smooth spatial transitions and photometric coherence.

Let \(P_{\text{FP}}, P_{\text{FN}},\) and \(P_{\text{Labeled}}\) denote the sets of image patches extracted from false positives, false negatives, and ground-truth labeled objects identified during prior training epochs. Formally, the augmentation process is defined as

{
\begin{equation}
I' \;=\; \mathrm{Aug}\!\Bigl(
   I,\;
   \{P_{\text{FPs}},\,P_{\text{FNs}},\,P_{\text{Labeled}}\}\;   ;\;
   c,\; \theta
\Bigr),
\end{equation}
}
\noindent where \(I\) is a background image, \(c\) the context map, and \(\theta\) the transformation parameters. The resulting image \(I'\) is a semantically augmented sample generated by embedding transformed patches into habitat-consistent regions of \(I\).

\begin{figure*}[t]
    \centering
    \includegraphics[width=\linewidth]{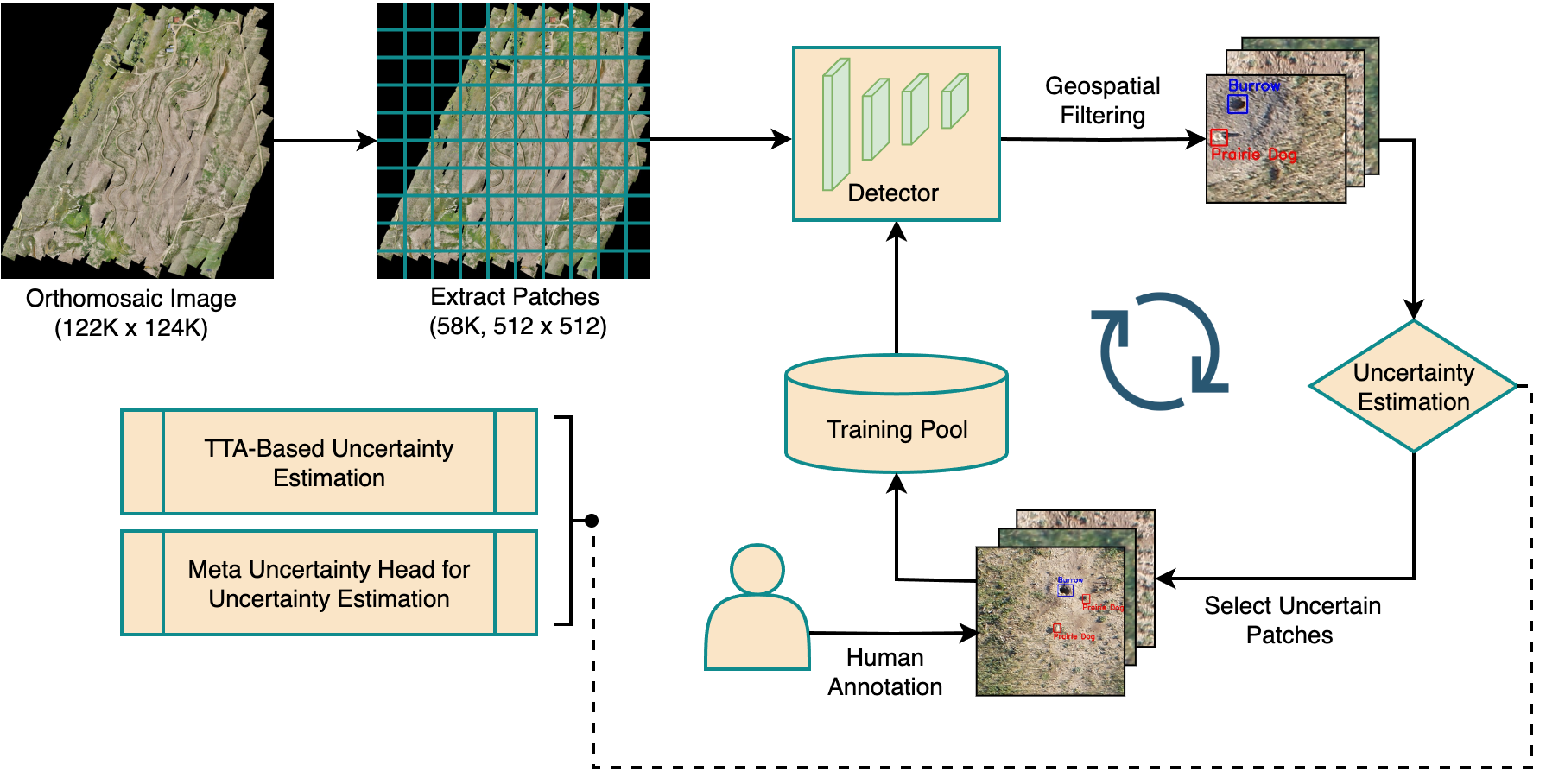} 
    \caption{Overview of the geospatial active learning pipeline. Large orthomosaic surveys are tiled into $512\times512$ patches. Following initial detection, spatial filtering and uncertainty estimation modules select the most informative tiles for expert annotation, forming a closed-loop training process.}
    \label{fig:activeLearning}
\end{figure*}

The patch sets \(P_{\text{FP}}, P_{\text{FN}},\) and \(P_{\text{Labeled}}\) capture common sources of detection ambiguity, including shadows, rocks, and vegetation. Their targeted reintegration into realistic backgrounds encourages the detector to learn more discriminative context cues for prairie dogs and burrows, improving robustness to visually similar clutter. Overall, the CA-HSA pipeline generated 3{,}929 additional training images, enriching the dataset with rare and ambiguous examples.

\section{RareSpot+ Geospatial Active Learning}
\label{sec:active_learning}

Inference over large aerial surveys comprising tens of thousands of tiles is computationally tractable; however, improving detector performance in such settings is fundamentally constrained by the cost of expert annotation, particularly when targets are rare and spatially sparse. Active learning (AL) addresses this challenge by querying the most informative unlabeled samples for annotation~\cite{Settles2009TR,Tuia2011TGRS,Brust2019ALOD}. Standard AL strategies, however, often assume sample independence and perform poorly when targets exhibit strong spatial clustering, as is common in ecological imagery~\cite{Legendre1993Ecology,Fortin2005Book}.

We introduce a geospatial active learning framework that explicitly exploits \emph{spatial relationships} between classes to guide sample acquisition. In the prairie dog monitoring setting, animal locations are strongly associated with nearby burrow entrances. Our geospatial analysis (Supplementary Materials, Section 3) shows that approximately 95\% of prairie dogs lie within 15.2\,m of a burrow, consistent with ecological observations~\cite{Hoogland1995Book}. We therefore use high-confidence burrow detections as a high-recall proxy to pre-filter the search space prior to uncertainty-based selection. While the specific spatial relation used here is domain-specific, the underlying principle—leveraging auxiliary spatial cues to guide annotation—is broadly applicable when such cues can be defined or learned.

The geospatial active learning pipeline is illustrated in Figure~\ref{fig:activeLearning}. An initial detector \(f_{\theta_0}\) is trained on a labeled dataset \(\mathcal{D}_{L}^{(0)}\). Large orthomosaics are then tiled into $512\times512$ patches, and inference is performed to identify a candidate set \(\mathcal{P}\) of tiles containing confident burrow detections. Tiles in \(\mathcal{P}\) are ranked using uncertainty measures \(U(x; f_{\theta_0})\) (Section~\ref{sec:acquisition_functions}), and the top-$k$ tiles (\(k \in \{100, 200, 500, 1000\}\)) are selected for expert annotation. The newly annotated tiles are merged with \(\mathcal{D}_{L}^{(0)}\) to retrain the detector, yielding an updated model \(f_{\theta_1}\). This relation-guided strategy focuses the annotation budget on geographically relevant and informative regions, substantially reducing redundant labeling in large-scale surveys.

\subsection{Uncertainty Estimation Models}
\label{sec:acquisition_functions}
We evaluate two tile-level uncertainty estimation strategies within the geospatial active learning pipeline shown in Figure~\ref{fig:activeLearning}.

\begin{figure}[h!]
    \centering
    \includegraphics[width=\linewidth]{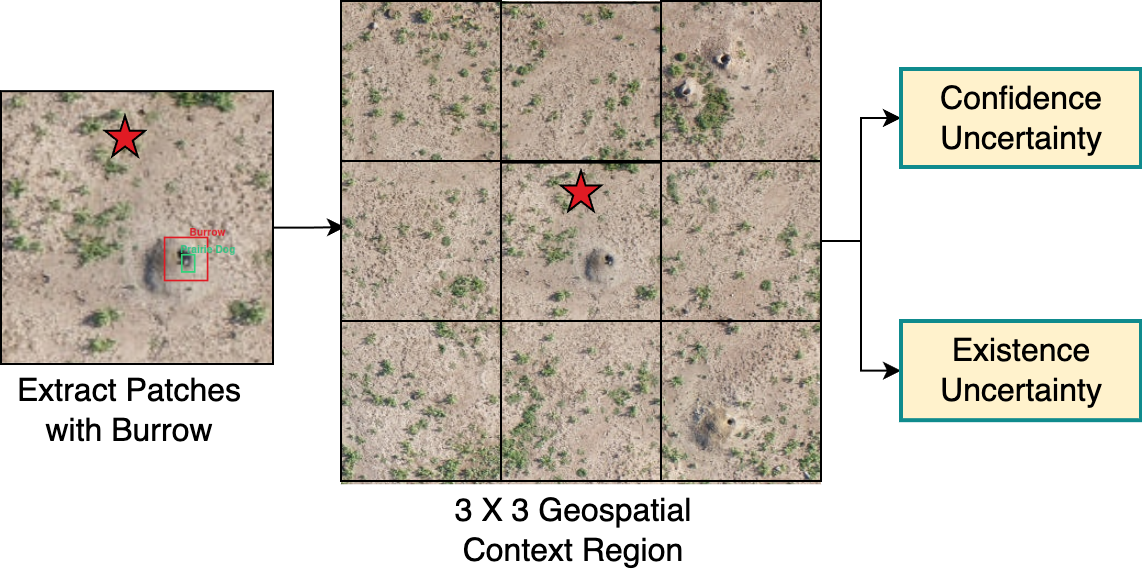}
    \caption{Sampling for TTA-based uncertainty.}
    \label{fig:tta}
\end{figure}

\subsubsection{TTA-Based Uncertainty with Geospatial Context}
\label{sec:model1_tta}
We estimate prediction uncertainty using instability under test-time augmentation (TTA), rather than stochastic inference techniques such as MC Dropout. For a candidate tile \(x\) and its $3\times3$ spatial neighborhood \(\mathcal{N}(x)\), we apply \(T\) augmentations \(a_t \sim \mathcal{A}\), with each augmentation applied identically across all nine tiles to preserve local spatial structure. This yields a set of predictions \(\{P_t\}_{t=1}^T\), as illustrated in Figure~\ref{fig:tta}. Detections across TTA passes are clustered using an IoU threshold \(\tau\) to form a set of unique object instances \(o \in \mathcal{O}_x\) (see Algorithm~1 in the Supplementary Materials).

\paragraph{Confidence Uncertainty (\(U_{\text{c}}\)).}
Let \(c_t(o) \in [0,1]\) denote the confidence score of instance \(o\) at augmentation \(t\), and let \(\bar c(o)=\tfrac{1}{T}\sum_t c_t(o)\) be the mean confidence. We define
\[
\sigma_c^2(o)=\frac{1}{T}\sum_{t=1}^{T}\big(c_t(o)-\bar c(o)\big)^2, \qquad
U_{\text{c}}(o)=4\,\sigma_c^2(o),
\]
which attains its maximum when predictions are split between high and low confidence values.

\paragraph{Existence Uncertainty (\(U_{\text{ex}}\)).}
Let \(k=T_{\text{detected}}(o)\) be the number of TTA passes in which instance \(o\) is detected. Using a Jeffreys prior, we estimate the detection probability as \(\hat p(o)=\tfrac{k+0.5}{T+1}\) and define
\[
U_{\text{ex}}(o)=4\,\hat p(o)\big(1-\hat p(o)\big),
\]
which corresponds to the normalized Bernoulli variance and peaks when object presence is ambiguous across TTA views.

\paragraph{Tile-Level Uncertainty Aggregation.}
Let \(\mathcal{O}_x^{\mathrm{pd}}\) and \(\mathcal{O}_x^{\mathrm{b}}\) denote the sets of prairie dog and burrow instances detected in tile \(x\). We normalize for instance count using
\[
\alpha_i(x)=\frac{1}{\max\!\big(1,|\mathcal{O}_x^{i}|\big)}, \qquad i \in \{\mathrm{pd}, \mathrm{b}\}.
\]
With class-specific weights \(w_{\text{c}}^{i}\) and \(w_{\text{ex}}^{i}\), the final tile-level uncertainty score is
\begin{equation}
\label{eq:tta_uncertainty}
\begin{aligned}
U(x) =
&\;\alpha_{\mathrm{pd}}(x)
\sum_{o \in \mathcal{O}_x^{\mathrm{pd}}}
\Big( w_{\text{c}}^{\mathrm{pd}} U_{\text{c}}(o)
    + w_{\text{ex}}^{\mathrm{pd}} U_{\text{ex}}(o) \Big) \\
+&\;\alpha_{\mathrm{b}}(x)
\sum_{o \in \mathcal{O}_x^{\mathrm{b}}}
\Big( w_{\text{c}}^{\mathrm{b}} U_{\text{c}}(o)
    + w_{\text{ex}}^{\mathrm{b}} U_{\text{ex}}(o) \Big).
\end{aligned}
\end{equation}
Prairie dog weights are typically set higher than burrow weights to emphasize the rarer and more annotation-critical class.

\subsubsection{Learned Meta-Uncertainty Head (MUH)}
\label{sec:model2_learned}
\begin{figure}[h!]
    \centering
    \includegraphics[width=\linewidth]{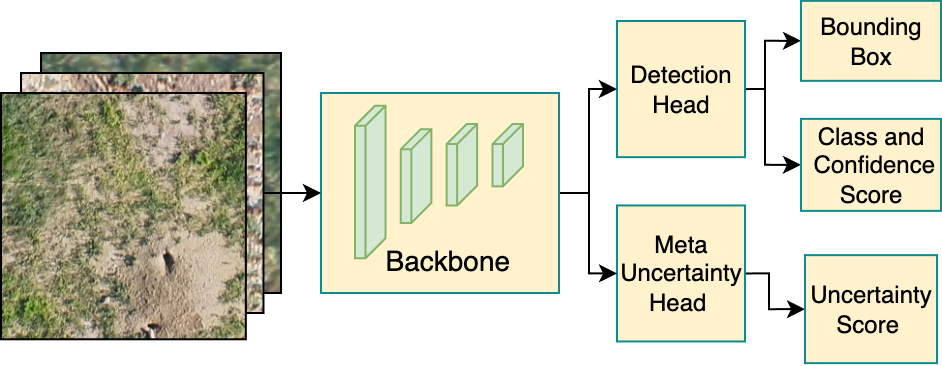}
    \caption{Architecture of the Meta-Uncertainty Head (MUH).}
    \label{fig:meta_learning}
\end{figure}

In addition to TTA-based uncertainty, we learn uncertainty through a lightweight \emph{Meta-Uncertainty Head} (MUH) that directly regresses a tile-level error score from internal detector features, treating uncertainty estimation as a meta-learning problem over the base detector~\cite{hospedales2021meta}. As illustrated in Figure~\ref{fig:meta_learning}, MUH branches from the frozen \(P_3\) feature map—chosen for its fine-grained spatial detail—of a detector trained with multi-scale consistency learning and context-aware augmentation, and outputs a single scalar uncertainty value per tile.

Supervision for MUH is provided by a self-labeled \emph{Prediction Uncertainty Score}, \(U_{\text{score}}\), computed on a held-out validation split with the base detector frozen. The score aggregates multiple sources of detection error, including localization inaccuracies, classification ambiguity, false negatives, and false positives:

\begin{align} \label{eq:uscore_aligned}
    U_{\text{score}} ={}& \frac{1}{|P| + |G| - |M|} \Biggl[ \nonumber \\
    &\sum_{(p_i, g_j) \in M} \bigl(1 - c_i \cdot \mathrm{IoU}(p_i, g_j)\bigr) \nonumber \\
    &+ |G_{FN}| + \sum_{p_i \in P_{FP}} \min(1, 0.5 + c_i) \Biggr].
\end{align}

\noindent
where \(P\) and \(G\) denote the sets of predictions and ground-truth objects, respectively, \(M\) is the set of matched prediction–ground-truth pairs, \(c_i\) is the confidence score of prediction \(p_i\), and \(\mathrm{IoU}(\cdot)\) measures localization overlap. The normalization term ensures that \(U_{\text{score}}\) is invariant to the number of objects per tile.

By training MUH to regress \(U_{\text{score}}\), the detector learns a direct mapping from its internal feature representations to expected failure modes. This enables a data-driven acquisition function that selects informative tiles based not only on prediction variance, but also on learned, task-specific indicators of detection difficulty.

\section{Experimental Evaluation}
\label{sec:experiments}

% We evaluate the proposed RareSpot framework to quantify the individual and combined impact of its core components—Multi-Scale Consistency Learning (MSCL) and Context-Aware Hard Sample Augmentation (CA-HSA)—on prairie dog and burrow detection performance. All experiments are conducted using identical training and evaluation protocols across the baseline and ablation variants to ensure fair comparison. The following subsections report quantitative and qualitative detection results, along with cross-dataset generalization to other drone-based wildlife benchmarks.

We evaluate the proposed RareSpot+ framework to quantify the individual and combined impact of its core components, including Multi-Scale Consistency Learning (MSCL), Context-Aware Hard Sample Augmentation (CA-HSA), and Geospatially-Guided Active Learning, on prairie dog and burrow detection performance. All experiments are conducted using identical training and evaluation protocols across the baseline and ablation variants to ensure fair comparison. The following subsections report quantitative and qualitative detection results, the efficiency of our uncertainty-driven acquisition strategies under varying annotation budgets, and cross-dataset generalization to other drone-based wildlife benchmarks.

\begin{table*}[t]

\centering
%charvi changes - correct grammer
\caption{RareSpot+ detection experiments and ablation study results showing the impact of different modules on both validation and test sets for prairie dog and burrow detection. The proposed \textit{RareSpot+}, which integrates Multi-Scale Consistency Learning and Context-Aware Hard Sample Augmentation, outperforms the baseline across both datasets. These results demonstrate the effectiveness of incorporating cross-scale feature consistency and hard sample prioritization to improve small-object detection.}

\label{tab:results}
\resizebox{\textwidth}{!}{
{
\begin{tabular}{l|cc|cc|cc|c}
\hline
Model & \multicolumn{2}{c|}{Precision (P)} & \multicolumn{2}{c|}{Recall (R)} & \multicolumn{2}{c|}{AP@50} & Overall mAP@50 \\
 & PD & Burrow & PD & Burrow & PD & Burrow & \\
\hline
\multicolumn{8}{c}{Validation Results} \\
\hline
Baseline Model (YOLOv5L) & 0.570 & 0.791 & 0.370 & 0.840 & 0.393 & 0.846 & 0.620 \\

Multi-Scale Consistency Learning & 0.578 & 0.824 & 0.434 & 0.850 & 0.432 & 0.843 & 0.637 \\
CA Hard Sample Augmentation & 0.557 & 0.823 & 0.407 & 0.851 & 0.415 & 0.852 & 0.633 \\
\textbf{RareSpot+ (Combined)} & \textbf{0.634} & \textbf{0.845} & \textbf{0.470} & \textbf{0.885} & \textbf{0.491} & \textbf{0.871} & \textbf{0.681} \\
\hline
Improvement over Baseline & \textbf{+11.23\%} & \textbf{+6.83\%} & \textbf{+27.03\%} & \textbf{+5.36\%} & \textbf{+24.94\%} & \textbf{+2.96\%} & \textbf{+9.90\%} \\
\hline
\multicolumn{8}{c}{Test Results} \\
\hline
Baseline Model (YOLOv5L) & 0.482 & 0.790 & 0.366 & 0.891 & 0.366 & 0.886 & 0.626 \\

Multi-Scale Consistency Learning & 0.536 & 0.842 & 0.488 & 0.910 & 0.454 & 0.914 & 0.684 \\
CA Hard Sample Augmentation & 0.464 & 0.871 & 0.471 & 0.891 & 0.383 & 0.912 & 0.647 \\
\textbf{RareSpot+ (Combined)} & \textbf{0.591} & \textbf{0.893} & \textbf{0.519} & \textbf{0.907} & \textbf{0.495} & \textbf{0.923} & \textbf{0.709} \\
\hline
Improvement over Baseline & \textbf{+22.61\%} & \textbf{+13.04\%} & \textbf{+41.80\%} & \textbf{+1.80\%} & \textbf{+35.25\%} & \textbf{+4.18\%} & \textbf{+13.26\%} \\
\hline
\end{tabular}
}
}
\end{table*}

\subsection{Detection Performance and Ablation Study}
To assess the individual and combined contributions of the proposed modules, we conduct a comprehensive ablation study, with results summarized in Table~\ref{tab:results}. The baseline model (YOLOv5-L with standard high-augmentation settings) achieves strong performance for burrow detection but performs less reliably on prairie dogs, which are smaller, visually subtle, and sparsely distributed.

\emph{Multi-Scale Consistency Learning} improves detection performance by reinforcing cross-scale feature alignment for small objects. On the validation set, this module improves prairie dog mAP@50 by 17.3\%, while on the test set, mAP@50 increases by 24\%, demonstrating its effectiveness in preserving fine-grained discriminative features critical for small-object detection.

\emph{Context-Aware Hard Sample Augmentation} further enhances robustness by emphasizing visually ambiguous instances through context-matched compositing of hard examples. This strategy improves both recall and precision for prairie dogs and burrows across datasets, indicating increased sensitivity to rare and challenging targets. In some settings, particularly when the training set is already large, the increased emphasis on hard negatives can lead to minor precision reductions, reflecting a common sensitivity–specificity trade-off in augmentation-based learning.

Finally, the full \emph{RareSpot+} detection model, which integrates both MSCL and CA-HSA, achieves the strongest overall performance. On the test set, RareSpot+ improves prairie dog mAP@50 by 35.25\% and recall by 41.80\%, demonstrating the complementary benefits of representation-level consistency and context-aware data augmentation for detecting small and infrequent objects in aerial imagery. All reported results are averaged over three independent runs with different random seeds; RareSpot+ achieves a mean mAP@50 of $0.681 \pm 0.009$, compared to $0.620 \pm 0.007$ for the baseline.

\begin{table}[t]
\centering
\caption{Comparison with other detection models on the validation set and parameter counts (M). All models are trained and evaluated under identical settings.}
\label{tab:othermodels}
\begin{tabular}{l|c|c}
\hline
\textbf{Models} & \textbf{mAP@50} & \textbf{Params} \\
\hline
Baseline (YOLOv5L) & 0.620 & 47M \\
YOLOv5L (P3 Only) & 0.623 & 46M \\
YOLOv7L & 0.517 & 71M\\
YOLOv8L & 0.597 & 43.7M\\
YOLOv10L & 0.471 & 26M\\
YOLOv11L & 0.589 & 26M\\
YOLOv12L & 0.609 & 26.4M\\
DETR & 0.411 & 41M\\
Co-DETR & 0.437 & 304M\\
TPH-YOLOv5 & 0.534 & 90M\\
WildlifeMapper & 0.313 & 636M\\
\textbf{RareSpot+} & \textbf{0.681} & 47M\\
\hline
\end{tabular}
\end{table}

\subsection{Choice of Backbone and Comparison with State-of-the-Art Methods}
We adopt YOLOv5 as the base detector due to its strong performance on small-object benchmarks and its ability to preserve fine-grained spatial detail through static feature aggregation and fixed anchor design. In comparative experiments (Tables~\ref{tab:results} and~\ref{tab:othermodels}), alternative FPN-based architectures—including ResNet-50-FPN, other YOLO variants, and ViT-FPN models—consistently underperformed relative to YOLOv5 on our dataset. Although more recent YOLO variants demonstrate improved performance on common benchmarks, we observe that YOLOv5-L remains particularly effective for extremely small, low-contrast targets in aerial imagery. We therefore focus on strengthening the most competitive baseline rather than modifying the underlying detection architecture.

While YOLOv7 shares a CSPDarknet–FPN backbone, its architectural modifications—such as RepConv, E-ELAN, and dynamic label assignment—are primarily optimized for medium-to-large objects and tend to reduce the fidelity of high-resolution features required for reliable small-object detection~\cite{li2024YOLOUAV,li2023improved}. In contrast, our Multi-Scale Consistency Learning module integrates seamlessly with YOLOv5, introduces no additional parameters, and requires no dataset-specific tuning, while consistently improving detection performance. These gains extend to mixed-scale datasets such as WAID, where object sizes range from 20 to 530 pixels.

As shown in Table~\ref{tab:othermodels}, the full RareSpot+ detection model outperforms state-of-the-art detectors, including transformer-based architectures (DETR~\cite{carion2020end}, Co-DETR~\cite{zong2023detrs}) and specialized small-object designs (TPH-YOLO~\cite{zhu2021tph}, WildlifeMapper~\cite{kumar2024wildlifemapper}). Our approach achieves superior performance while maintaining a substantially lower parameter count than these specialized models. Moreover, many competing methods rely on extensive annotation budgets and large-scale supervision, which limits their effectiveness in data-sparse, rare-species monitoring scenarios. In contrast, our model is designed to operate efficiently under limited annotation budgets and was trained in approximately four hours on four NVIDIA A100 GPUs using a batch size of 64.

\subsection{RareSpot+ Detection Performance on Other Wildlife Datasets}
To assess the broader applicability of the RareSpot+ framework, we evaluate its detection and augmentation components on five diverse drone-based wildlife datasets: the Aerial Elephant Dataset (AED)~\cite{Naude_2019_CVPR_Workshops}, the Waterfowl Thermal Imagery Dataset~\cite{Huwaterfowl2024}, WAID~\cite{mou2023waid}, HerdNet~\cite{Delplanque2023ISPRS}, and the dataset of Eikelboom \emph{et al.}~\cite{eikelboom2019improving}. These experiments are designed to evaluate the portability of the RareSpot+ detection architecture—specifically Multi-Scale Consistency Learning (MSCL) and Context-Aware Hard Sample Augmentation (CA-HSA)—rather than the full geospatial active learning framework.

Unlike our primary prairie dog study, these benchmarks feature larger targets (approximately 45–166 pixels in width) and do not provide explicit geospatial habitat proxies. As such, they do not exhibit the extreme small-object scale or strongly clustered spatial distributions that motivate geospatial active learning. Results on these datasets therefore reflect detector-level portability rather than full-system generalization.

Despite differences in sensing modality (visible versus thermal imagery) and target species, RareSpot+ consistently outperforms the YOLOv5 baseline across multiple benchmarks. As shown in Table~\ref{tab:otherdatasets}, MSCL yields consistent performance gains on datasets with smaller object instances, achieving 97.6\% mAP@50 on WAID and outperforming both WildlifeMapper and the standard YOLOv5 architecture. To further examine component-wise generalization, Table~\ref{tab:otherdatasets_2} reports results for MSCL and CA-HSA individually on the HerdNet and Eikelboom datasets. 
%In the absence of explicit metadata relating targets to their ecological context in these benchmarks, we applied a strategy assuming a higher frequency of animal presence in exposed soil, consistent with the methodology used for our prairie dog dataset. 
In the absence of explicit metadata relating targets to their ecological context in these benchmarks, we employed a simple, dataset-agnostic foreground–background heuristic for context-aware sample placement, rather than the domain-specific geospatial priors used in the prairie dog study.
Visualizations of the augmentation results for the Eikelboom et al. and HerdNet datasets are provided in Figure 2 of the Supplementary Materials. Although these benchmarks contain larger targets and different ecological distributions, both components independently improve upon the baseline, with their combination in the full \textbf{RareSpot+} model achieving the strongest overall performance.

\begin{table}[t]
{
    
    \caption{Detection performance (mAP@50) on other drone-based wildlife datasets. Results compare a YOLOv5 baseline, WildlifeMapper (WM), and YOLOv5 augmented with Multi-Scale Consistency Learning (MSCL). These experiments evaluate detector-level portability rather than the full RareSpot+ framework.}
    \centering
    \begin{tabular}{lccc}
        \hline
        Method & AED & Waterfowl Thermal & WAID\\
        \hline
        YOLOv5 & 0.891 & 0.957 & 0.956\\
        WM & 0.553 & 0.368 & 0.658\\
      %  SE-YOLO & & & 0.983 \\
        \textbf{MSCL} & \textbf{0.897} & \textbf{0.970} & \textbf{0.976} \\
        \hline
    \end{tabular}
    
    \label{tab:otherdatasets}
}
\end{table}

\begin{table}[ht]
\caption{Detection performance (mAP@50) on the HerdNet and Eikelboom wildlife datasets. Results compare the YOLOv5-L baseline with individual RareSpot+ components—Multi-Scale Consistency Learning (MSCL) and Context-Aware Hard Sample Augmentation (CA-HSA)—as well as their combination. The “Original Paper” row reports results as published by the respective dataset authors.}

\centering
\setlength{\tabcolsep}{8pt} % Slightly increased for better readability
\renewcommand{\arraystretch}{1.2} % Adds a bit of breathing room to rows
\begin{tabular}{lcc}
\hline
Model & Herdnet & Eikelboom \\
\hline
Original Paper & 0.800 & 0.810 \\
Baseline (YOLOv5L) & 0.773 & 0.830 \\
MSCL & 0.848 & 0.836 \\
CA-HSA & 0.798 & 0.842 \\
\textbf{MSCL + CA-HSA} & \textbf{0.870} & \textbf{0.852} \\ % Added a 0 to 0.870 for decimal consistency
\hline
\end{tabular}
\label{tab:otherdatasets_2}
\end{table}

\subsection{Qualitative Analysis}
Qualitative visualizations (Figure~\ref{fig:feature_map}; see also Section~6 of the Supplementary Materials ) show that RareSpot+ localizes prairie dogs and burrows more reliably than the baseline and produces cleaner, target-focused feature activations. For visualization purposes, high-dimensional feature maps are aggregated by averaging activations across the channel dimension and normalized to produce a single intensity map; this procedure is used exclusively for qualitative analysis.

These activation maps illustrate the effect of \textit{multi-scale consistency learning}: by explicitly aligning adjacent detection heads ($P_3{\leftrightarrow}P_4$ and $P_4{\leftrightarrow}P_5$), the model improves cross-scale feature coherence while preserving the fine spatial detail required for reliable small-object localization. As a result, RareSpot+ yields more localized and less noisy activations around true targets compared to the baseline. Additional qualitative examples and visual analyses are provided in Section~6 of the Supplementary Materials.

In summary, RareSpot+ achieves substantial performance gains over strong baselines through its unified treatment of scale alignment and context-aware augmentation. These improvements yield more stable and interpretable detection outputs, which in turn support downstream spatial analysis and uncertainty-aware annotation strategies evaluated in the following sections.

% AL
\begin{table*}[ht]
\centering
\caption{Active learning performance vs. annotation budget. Bold indicates the best metric at each budget across strategies. PD: Prairie Dog.}
\label{tab:al_results_detailed}
\resizebox{\textwidth}{!}{%
\begin{tabular}{l c cc cc cc c}
\toprule
\multirow{2}{*}{\textbf{Acquisition Strategy}} & \multirow{2}{*}{\textbf{Budget (\textit{k})}} & \multicolumn{2}{c}{\textbf{Precision (P)}} & \multicolumn{2}{c}{\textbf{Recall (R)}} & \multicolumn{2}{c}{\textbf{AP@50}} & \multirow{2}{*}{\textbf{Overall mAP@50}} \\
\cmidrule(lr){3-4} \cmidrule(lr){5-6} \cmidrule(lr){7-8}
& \textbf{Out of 58k} & \textbf{PD} & \textbf{Burrow} & \textbf{PD} & \textbf{Burrow} & \textbf{PD} & \textbf{Burrow} & \\
\midrule
Initial Model (RareSpot+) & 0 & 0.591 & 0.893 & 0.519 & 0.907 & 0.495 & 0.923 & 0.709 \\
\midrule
\multirow{3}{*}{Random Sampling} 
& 100  & 0.558 & 0.897 & 0.523 & 0.902 & 0.485 & 0.922 & 0.704 \\
& 500  & 0.580 & 0.898 & 0.539 & 0.909 & 0.510 & 0.925 & 0.717 \\
& 1000 & 0.572 & 0.897 & 0.539 & 0.911 & 0.491 & 0.928 & 0.709 \\
% \midrule
% \multirow{3}{*}{MC Dropout} 
% & 100  & 0.525 & 0.815 & 0.460 & 0.901 & 0.431 & 0.895 & 0.663 \\
% & 500  & 0.548 & 0.829 & 0.505 & 0.908 & 0.472 & 0.902 & 0.687 \\
% & 1000 & 0.561 & 0.840 & 0.532 & 0.911 & 0.498 & 0.905 & 0.702 \\
\midrule
\multirow{3}{*}{\begin{tabular}{@{}l@{}}Geospatial + \\ Random Sampling\end{tabular}} 
& 100  & 0.583 & 0.911 & 0.512 & 0.904 & 0.498 & 0.928 & 0.713 \\
& 500  & 0.599 & 0.909 & 0.535 & 0.907 & 0.515 & 0.928 & 0.722 \\
& 1000 & 0.563 & 0.893 & 0.568 & 0.919 & 0.508 & 0.931 & 0.719 \\
\midrule
\multirow{3}{*}{\begin{tabular}{@{}l@{}}Geospatial + \\ TTA (Ours)\end{tabular}} 
& 100  & \textbf{0.617} & 0.893 & 0.510 & \textbf{0.916} & \textbf{0.516} & \textbf{0.927} & \textbf{0.721} \\
& 500  & \textbf{0.627} & 0.903 & 0.501 & \textbf{0.921} & 0.516 & \textbf{0.934} & 0.725 \\
& 1000 & \textbf{0.589} & 0.896 & 0.553 & \textbf{0.924} & 0.536 & \textbf{0.936} & 0.736 \\
\midrule
\multirow{3}{*}{\begin{tabular}{@{}l@{}}\textbf{Geospatial +} \\ \textbf{Meta-Uncertainty Head (Ours)}\end{tabular}} 
& 100  & 0.551 & 0.892 & \textbf{0.546} & 0.907 & 0.510 & 0.921 & 0.716 \\
& 500  & 0.578 & \textbf{0.908} & \textbf{0.561} & 0.904 & \textbf{0.533} & 0.930 & \textbf{0.732} \\
& 1000 & 0.566 & \textbf{0.903} & \textbf{0.571} & 0.905 & \textbf{0.548} & 0.925 & \textbf{0.737} \\
\bottomrule
\end{tabular}
}
\end{table*}

\subsection{Active Learning Results and Analysis}
\label{sec:al_results}
Table~\ref{tab:al_results_detailed} summarizes active learning performance as a function of annotation budget. \emph{Random Sampling} yields only marginal improvements even at 1{,}000 annotated tiles, reflecting the difficulty of identifying informative samples when targets are rare and spatially sparse. The maximum budget of 1,000 tiles was selected to reflect the practical constraints of expert availability and the significant labor costs associated with high-precision ecological annotation. Incorporating a geospatial prior (\emph{Geospatial + Random Sampling}) substantially improves outcomes by restricting candidate selection to burrow-adjacent tiles; for example, prairie dog recall reaches 0.568 at 1{,}000 tiles compared to 0.539 under random sampling.

Within this focused candidate pool, uncertainty-based acquisition becomes critical. Both \emph{Geospatial + TTA (Ours)} and \emph{Geospatial + Meta-Uncertainty Head (Ours)} outperform geospatial-only selection, with the learned MUH achieving the strongest overall performance. With 1{,}000 annotated tiles (approximately 1.7\% of the 58k-tile survey), MUH attains the highest overall mAP@50 (0.737) and delivers the largest gains on the rare class: prairie dog recall increases from 0.519 to 0.571, and AP improves from 0.495 to 0.548.

The two uncertainty strategies exhibit complementary behavior. TTA-based uncertainty measures prediction instability across augmentations, which tends to surface borderline confounders such as shadows or complex vegetation. This effectively supports targeted hard negative mining, leading to a primary improvement in precision (e.g., +2.6\% at $k=100$) as the detector becomes more conservative. In contrast, the Meta-Uncertainty Head is trained to predict a task-specific failure score (\(U_{\text{score}}\)) that explicitly emphasizes expected false negatives. This \emph{failure discovery} mechanism drives stronger recall improvements (e.g., +2.7\% at $k=100$) by prioritizing subtle, low-contrast prairie dogs that are often missed by the base detector.

Across all strategies, most of the performance gains for prairie dogs arise from increased recall rather than precision. This reflects the core effect of the RareSpot+ design: Multi-Scale Consistency Learning and Context-Aware Hard Sample Augmentation primarily recover subtle, low-contrast instances that the baseline fails to detect, thereby reducing false negatives. Similarly, the geospatially guided active learning pipeline concentrates annotation effort on burrow-adjacent regions where prairie dogs are likely but underrepresented. While spatial priors reduce obvious background false positives and help stabilize precision relative to random sampling, some visual confounders remain. In ecological monitoring, such recall-oriented gains are particularly desirable, as minimizing missed detections of rare individuals is often more critical than eliminating occasional false alarms.

\begin{figure*}[p]
    \centering
    \includegraphics[width=\linewidth]{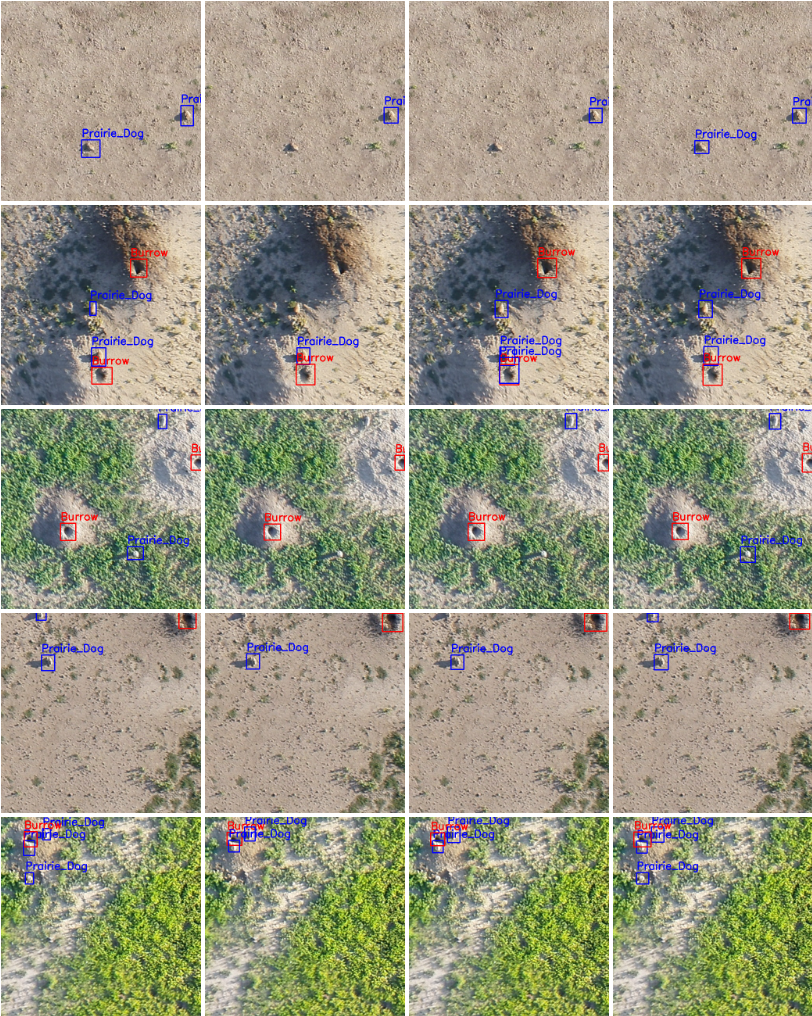}
      \caption{Active Learning qualitative results. \textbf{Left$\to$Right:} Ground truth, Baseline YOLOv5, RareSpot+, and RareSpot+ after AL. The AL iteration reduces both false negatives and false positives and sharpens localization.}
\end{figure*}

\section{Discussion, Geospatial Generalization, and Future Directions}
\label{sec:discussion}

\subsection{Summary of Contributions}
This work addresses two persistent challenges in aerial wildlife monitoring: reliable detection of small, rare targets and the high cost of expert annotation. The proposed \textbf{RareSpot+} framework demonstrates that scalable, data-efficient monitoring can be achieved by jointly advancing representation learning, data augmentation, and annotation strategy. Multi-Scale Consistency Learning preserves fine-grained spatial features critical for small-object localization, while Context-Aware Hard Sample Augmentation improves robustness to visually confounding backgrounds, together yielding a 35\% improvement in mAP@50 over a strong baseline. To further reduce labeling effort, the Geospatial Active Learning framework integrates ecological priors with uncertainty-aware acquisition, achieving a 14.5\% improvement in prairie dog average precision using only 1.7\% of the total survey tiles. %Collectively, these components form an end-to-end system that scales effectively to large aerial datasets and supports downstream spatial analyses such as colony-level structure modeling.
Collectively, these components form a complete system that scales effectively to large aerial surveys and supports downstream spatial analyses such as colony-level structure modeling.

\subsection{Geospatial Generalization and Ecological Implications}
While the specific geospatial relation exploited in this study—the co-occurrence of prairie dogs and burrows—is domain-specific, RareSpot+ establishes a broader methodological principle for ecological monitoring: the explicit integration of auxiliary environmental cues to guide sample acquisition for rare targets. In natural ecosystems, species distributions are rarely random; instead, they are closely coupled to habitat structures and persistent environmental markers. RareSpot+ provides a blueprint for incorporating such species–habitat spatial priors within an active learning framework to improve annotation efficiency.

This paradigm is applicable to a range of ecological monitoring scenarios in which auxiliary spatial cues are available. For example, many colonial or burrowing mammals (e.g., meerkats and other rodents) exhibit strong dependence on persistent burrow systems that can serve as spatial proxies for animal presence. Large herbivores such as bison create wallows—distinct ground depressions that act as long-lived indicators of activity even when animals are not directly observable. In avian monitoring, seabird nesting colonies and rookeries provide fixed spatial anchors that enable targeted high-resolution surveys during breeding seasons. Although many existing wildlife datasets do not yet include the dual annotations (animal and habitat proxy) required to directly evaluate such relations, our prairie dog results serve as a proof of concept for the proposed approach.

\subsection{Boundary Conditions and Future Directions}
We emphasize that the geospatial active learning component is applicable only when meaningful auxiliary spatial cues can be defined or learned. In scenarios where such cues are absent, RareSpot+ naturally reduces to its detection and augmentation components, which remain broadly applicable across domains and datasets. This explicit boundary condition ensures that the framework does not rely on assumptions that cannot be met in practice.

Several directions for future work remain. Semi-supervised and self-supervised extensions could exploit large pools of unlabeled aerial imagery, while human-in-the-loop adaptation may enable continuous refinement in new environments. Incorporating temporal cues or multi-sensor data, such as thermal imagery or LiDAR, could further improve detection robustness and uncertainty calibration. More broadly, the principles demonstrated here—multi-scale feature alignment, context-aware augmentation, and relation-guided active learning—extend beyond ecology to other domains where small or rare instances are critical, including search and rescue, industrial inspection, and medical imaging. Because the proposed regularization is architectural-agnostic and applied only during training, it can be retrofitted to existing detection pipelines without increasing inference cost, a key property for large-scale deployments under real-world constraints.

Overall, RareSpot+ provides a unified and practical framework for small-object detection and data-efficient learning, establishing a foundation for scalable, automated analysis of fine-grained patterns in complex aerial imagery.

\section{Acknowledgement}
This research was supported in part by the NSF CSSI Award \#2411453. We thank the UCSB BisQue team: Chandrakanth Gudavalli, Connor Leverson, Amil Khan, and Umang Garg for their technical support and Lacey Hughey and Jared Stabach at the Smithsonian's National Zoo \& Conservation Biology Institute. We thank the annotation volunteers Spencer Harman, Lani O'Foran, Autumn Gray, Sydney Houck, Jessica Winey, Emily Csizmadia, and Isabella Barrera for improving our dataset quality.  We also extend our gratitude to American Prairie, especially Danny Kinka and Dan Stevenson, for their support and collaboration on this project.

\section*{Conflict of interest}
The authors declare that they have no known competing financial interests or personal relationships that could have appeared to influence the work reported in this paper.

\section*{Compliance with ethical standards}
This study did not involve human participants, human data, or human tissue. Aerial wildlife surveys were conducted in accordance with applicable animal welfare and wildlife-protection regulations. No animals were captured, handled, or disturbed for this research. Flight operations complied with relevant aviation regulations (e.g., FAA Part~107 in the United States), and land-access permissions were obtained from landowners and managing agencies prior to data collection. 

\section*{Informed consent}
Not applicable. This research did not involve human participants or identifiable personal data.

\section*{Funding}
This research was supported in part by the NSF CSSI Award \#2411453. The funders had no role in study design, data collection and analysis, decision to publish, or preparation of the manuscript.

\section*{Data availability}
% The data and code will be available at the project's repository: \url{https://github.com/eebowen/RareSpot}.
The data and code will be made available through Bisque UCSB~\cite{rarespot_dataset}.
% \href{https://bisque2.ece.ucsb.edu/client_service/view?resource=https://bisque2.ece.ucsb.edu/data_service/00-B2mLaCeFPkCgZF4LND9uEd}{Bisque UCSB}.
\clearpage
% % \bibliographystyle{./bst/sn-nature}
% % \bibliographystyle{sn-basic} 
% \bibliographystyle{apacite}
% \bibliography{sn-bibliography}% common bib file

%%===========================================================================================%%
%% If you are submitting to one of the Nature Portfolio journals, using the eJP submission   %%
%% system, please include the references within the manuscript file itself. You may do this  %%
%% by copying the reference list from your .bbl file, paste it into the main manuscript .tex %%
%% file, and delete the associated \verb+\bibliography+ commands.                            %%
%%===========================================================================================%%

% \bibliography{sn-bibliography}% common bib file
%% if required, the content of .bbl file can be included here once bbl is generated
%%\input sn-article.bbl

\end{document}